\documentclass{article}

\usepackage{microtype}
\usepackage{graphicx}
\usepackage{subcaption}
\usepackage{booktabs} 

\usepackage{hyperref}

\usepackage[preprint]{icml2026}

\usepackage{amsmath}
\usepackage{amssymb}
\usepackage{mathtools}
\usepackage{amsthm}

\usepackage{url}

\usepackage{algorithm}
\usepackage{algorithmic}

\usepackage{booktabs}

\usepackage{booktabs}
\usepackage{multirow}

\usepackage{graphicx}

\newcommand{\I}{\mathbf{I}}

\newcommand{\U}{\mathbf{U}}
\newcommand{\V}{\mathbf{V}}
\newcommand{\W}{\mathbf{W}}

\newcommand{\bfLambda}{\mathbf{\Lambda}}

\usepackage[capitalize,noabbrev]{cleveref}

\theoremstyle{plain}

\theoremstyle{definition}

\theoremstyle{remark}

\usepackage[textsize=tiny]{todonotes}

\icmltitlerunning{Scalable Variational Bayesian Fine-Tuning of LLMs via Orthogonalized Low-Rank Adapters}

\begin{document}

\twocolumn[
  \icmltitle{Scalable Variational Bayesian Fine-Tuning of LLMs \\via Orthogonalized Low-Rank Adapters}



  \icmlsetsymbol{equal}{*}

  \begin{icmlauthorlist}

  \icmlauthor{Haotian Xiang}{uga}
  \icmlauthor{Bingcong Li}{eth}
  \icmlauthor{Qin Lu}{uga}

  \end{icmlauthorlist}


  \icmlaffiliation{uga}{School of Electrical and Computer Engineering, University of Georgia, Athens, GA, USA}
\icmlaffiliation{eth}{Department of Computer Science, ETH Zurich, Zurich, Switzerland}

\icmlcorrespondingauthor{Qin Lu}{qin.lu@uga.edu}

  \icmlkeywords{Machine Learning, ICML}

  \vskip 0.3in
]



\printAffiliationsAndNotice{}  

\begin{abstract}
  When deploying large language models (LLMs) to safety-critical applications, uncertainty quantification (UQ) is of utmost importance to self-assess the reliability of the LLM-based decisions. However, such decisions typically suffer from overconfidence, particularly after parameter-efficient fine-tuning (PEFT) for downstream domain-specific tasks with limited data. 
Existing methods to alleviate this issue either rely on Laplace approximation based post-hoc framework, which may yield suboptimal calibration depending on the training trajectory, or variational Bayesian training that requires multiple complete forward passes through the entire LLM backbone at inference time for Monte Carlo estimation, posing scalability challenges for deployment.
To address these limitations, 
we build on the Bayesian last layer (BLL) model, where the LLM-based {\it deterministic} feature extractor is followed by random LL parameters for uncertainty reasoning. 
Since existing low-rank adapters (LoRA) for PEFT have limited expressiveness due to rank collapse, we address this with Polar-decomposed Low-rank Adapter Representation (PoLAR),
an orthogonalized parameterization paired with Riemannian optimization to enable more stable and expressive adaptation.
Building on this PoLAR-BLL model, we leverage the variational (V) inference framework to put forth a scalable Bayesian fine-tuning approach which jointly seeks the PoLAR parameters and approximate posterior of the LL parameters via alternating optimization.
The resulting PoLAR-VBLL is a flexible framework that nicely integrates architecture-enhanced optimization with scalable Bayesian inference to endow LLMs with well-calibrated UQ.
Our empirical results verify the effectiveness of PoLAR-VBLL in terms of generalization and uncertainty estimation on both in-distribution and out-of-distribution data for various common-sense reasoning tasks.  
\end{abstract}

\section{Introduction}

Large language models (LLMs) have demonstrated remarkable capabilities across diverse domains, from natural language understanding to complex reasoning tasks \citep{brown2020language, touvron2023llama}. When deploying to safety-critical applications, uncertainty quantification (UQ) is of utmost importance to self-assess the reliability of the LLM-based decisions. While large-scale pre-trained models exhibit reasonable calibration during pre-training \citep{kadavath2022language}, they fail to accurately express predictive uncertainty after parameter-efficient fine-tuning (PEFT) using limited data in downstream tasks~\citep{jiang2021calibrating}. Particularly, fine-tuned LLMs often exhibit significant overconfidence, which poses serious risks in high-stakes scenarios where reliable uncertainty estimation is essential for trustworthy decision-making~\citep{yang2023bayesian}.

To endow fine-tuned LLMs with well-calibrated UQ, several attempts have been made by leveraging advances in Bayesian neural networks (BNNs). Ensemble approaches \citep{lakshminarayanan2017simple, wang2023lora} require training multiple model copies, which incurs significant computational overhead~\citep{wang2023lora}. 
Post-hoc methods like Laplace approximation (LA)~\citep{yang2023bayesian} apply Bayesian inference after MAP estimation. However, LA captures only local uncertainty around a single mode, and its effectiveness may be limited when deterministic training converges to a region less amenable to uncertainty estimation~\citep{laplace2021,eschenhagen2021mixtures} (cf. Table~\ref{tab:main_results_BLOB_10_Only} for empirical evidence).
Variational methods like BLoB \citep{wang2024blob} pioneered joint optimization of mean and covariance during training for Bayesian PEFT. Despite its theoretical elegance, BLoB requires  multiple  forward passes through the entire LLM backbone during inference for Monte Carlo estimation, which poses scalability challenges for latency-sensitive applications. Subsequent variants such as ScalaBL~\citep{samplawski2025scalable}, C-LoRA~\citep{rahmati2025c}, and TFB~\citep{shi2024training} have made notable progress in reducing memory and computational costs, though efficient variational Bayesian fine-tuning that avoids multiple backbone forward passes remains an open challenge.
Going beyond these approaches, there are other methods for UQ in BNNs, including deep kernel learning~\citep{wilson2016deep} and variational Bayesian last layers (VBLL)~\citep{harrison2024variational}, which have not been explored for LLM fine-tuning.


On the other hand, the prohibitive computational cost of full fine-tuning has motivated the widespread adoption of PEFT methods. However, recent work on uncertainty quantification has established that Bayesian last layer (BLL) methods critically depend on the geometric properties of learned features~\citep{liu2020simple, postels2021practicality}. Specifically, \emph{distance-aware features}, where semantically distinct inputs remain well-separated in feature space, are essential for the BLL to reliably distinguish in-distribution from out-of-distribution samples.
This requirement poses a fundamental challenge for standard Low-Rank Adaptation (LoRA)~\citep{hu2022lora}, which suffers from directional diversity collapse where the stable rank  (a smooth proxy for effective dimensionality;  see Eq.~\eqref{eq:stable_rank} in App.~\ref{sec:stable_rank_analysis} for more details) often approaches 1, severely underutilizing the allocated subspace~\citep{zhang2025polar}. Such geometric compression directly undermines the distance-awareness requirement, projecting diverse inputs onto a narrow subspace and limiting the effectiveness of downstream Bayesian inference. Alternative approaches like DoRA~\citep{liu2024dora} and AdaLoRA~\citep{zhang2023adalora} have attempted to improve LoRA but still exhibit suboptimal rank utilization. The recently proposed PoLAR~\citep{zhang2025polar} addresses these limitations through polar decomposition with orthogonality constraints, preserving multi-directional feature geometry essential for UQ. However, most existing Bayesian LLM fine-tuning approaches still adopt vanilla LoRA, leaving the potential of architecture-aware optimization for UQ largely unexplored.

To bridge this gap, we propose PoLAR-VBLL, a principled framework in which each component addresses a specific challenge that the others cannot. The contributions of this paper are summarized as follows.
\begin{itemize}
    \item Relying on the BLL model, where the LLM-based {\it deterministic} feature extractor is followed by random LL parameters for UQ, we leverage the PoLAR-based LLM adapter~\citep{zhang2025polar}, an orthogonalized parameterization that preserves multi-directional feature geometry essential for distance-aware uncertainty estimation~\citep{liu2020simple}. This addresses the rank collapse in standard LoRA, which undermines downstream Bayesian inference. Our stable rank analysis (cf. App.~\ref{sec:stable_rank_analysis}) empirically validates this geometric preservation.  
\item The resulting PoLAR-BLL model is amenable to variational (V) training, where we jointly seek the PoLAR parameters via efficient landing field methods in Riemannian optimization and the approximate posterior of the LL parameters. We optimize a closed-form Jensen-tightened evidence lower bound (ELBO), and our ablation confirms that it remains tight throughout training (cf. Table~\ref{tab:jensen_vs_mc}, App.~\ref{sec:jensen_bound_tightness}). Since uncertainty is confined to the last layer, inference only requires a single backbone pass followed by multiple lightweight LL evaluations, whereas sampling-based methods~\citep{wang2024blob, shi2024training, samplawski2025scalable, rahmati2025c} require multiple complete LLM backbone forward passes, yielding a significant speedup (cf. Table~\ref{tab:uncertainty_ablation_GPU}, App.~\ref{sec: Memory usage and Run time}). 
    \item Further, we apply post-hoc LA~\citep{yang2023bayesian} to refine the variational covariance using exact Hessian information. The key insight is that VBLL discovers a favorable posterior mode, providing LA with a superior initialization. This is in contrast to standard pipelines that apply LA directly after deterministic training (Table~\ref{tab:main_results_BLOB_10_Only}). Notably, VBLL alone already achieves strong calibration and LA serves as a further refinement.  
    \item Comprehensive evaluations on \texttt{LLaMA-3.1-8B} and \texttt{LLaMA-2-7B} across six benchmarks demonstrate that PoLAR-VBLL consistently outperforms existing approaches in both accuracy and uncertainty calibration on in-distribution and out-of-distribution tasks (see Table~\ref{tab:main_results_BLOB_10_Only} and Table~\ref{tab:main_results_full} in App.~\ref{sec: Full comparison}).
\end{itemize}

\section{Related Work}

\subsection{UQ for Fine-Tuned LLMs and BNNs}

While large-scale pre-trained models exhibit reasonable calibration during pre-training \citep{kadavath2022language}, they fail to accurately express predictive uncertainty after fine-tuning \citep{jiang2021calibrating}, particularly when adapted to domain-specific tasks with limited data. This degradation necessitates Bayesian approaches for reliable uncertainty estimation in safety-critical applications.
Recent Bayesian PEFT methods exhibit limitations. Ensemble approaches \citep{lakshminarayanan2017simple} require training multiple LoRA copies with significant computational overhead. Laplace-LoRA \citep{yang2023bayesian} applies post-hoc approximation after MAP estimation, but this bifurcated optimization leads to suboptimal posterior estimates. 
BLoB \citep{wang2024blob} pioneered variational inference directly on LoRA parameters during training, achieving joint mean-covariance optimization. Despite its theoretical elegance, BLoB requires multiple forward passes through the entire LLM backbone at inference for Monte Carlo estimation, posing scalability challenges for latency-sensitive applications, while remaining fundamentally constrained by LoRA's low stable rank.
Several variants aim to reduce BLoB’s demanding memory cost. ScalaBL~\citep{samplawski2025scalable} uses stochastic subspace inference to reduce the number of variational parameters; C-LoRA~\citep{rahmati2025c} replaces them with deterministic contextual MLPs; and TFB~\citep{shi2024training} applies post-hoc search for Bayesian inference of deterministically trained adapters. These methods still rely on LLM backbone Monte Carlo sampling, posing challenges for inference efficiency.
VBLL \citep{harrison2024variational} demonstrates superior computational efficiency compared to full-adapter Bayesian methods by only considering the uncertainty of the last layer. 
During training, it leverages a closed-form Jensen-tightened ELBO, avoiding Monte Carlo sampling through the backbone. In inference, only a single backbone pass is needed, followed by multiple lightweight last-layer evaluations.
Previous applications in Bayesian optimization demonstrate its effectiveness \citep{brunzema2024bayesian}, but its adaptation to LLM fine-tuning with advanced adapter architectures, such as PoLAR, remains unexplored.

\subsection{PEFT}

The prohibitive computational cost of full fine-tuning for billion-parameter LLMs has made PEFT essential. LoRA \citep{hu2022lora} has gained widespread adoption by learning additive low-rank updates $\Delta \W$ on top of the frozen pre-trained weights $\W$.
Subsequent work has aimed to improve LoRA's effectiveness further. AdaLoRA \citep{zhang2023adalora} introduces adaptive rank allocation during training. DoRA \citep{liu2024dora} decomposes weights into magnitude and direction components. GaLore \citep{zhao2024galore} applies low-rank projection to optimizer states to reduce memory requirements.
However, recent analysis reveals fundamental limitations: LoRA suffers from directional diversity collapse where the stable rank of $\Delta \W$ remains well below the allocated linear algebraic rank, limiting expressiveness.
PoLAR addresses this through a re-parametrization with orthogonal constraints on direction matrices, and a tailored Riemannian optimization \citep{ablin2022fast} is employed for faster training on GPUs. In spite of these advances, adaptation to the Bayesian counterparts remains a rather uncharted territory -- existing Bayesian fine-tuning approaches all rely on the vanilla LoRA. 


\section{Variational Training of LLM-based Bayesian Last Layer Model via Orthogonal Low-Rank Adaptation}
Toward adapting the advances of PEFT so as to endow uncertainty-aware fine-tuning of LLMs with scalability, we present a unified framework that combines Polar-decomposed Low-rank Adapter Representation (PoLAR) with Variational Bayesian Last Layers (VBLL). The resulting approach addresses the fundamental limitation of standard LoRA's low stable rank while providing calibrated uncertainty estimates through scalable variational Bayesian inference.

\subsection{Bayesian last layer model with LLM-based feature extractor}

To endow LLM-based inference with UQ, we will rely on the Bayesian last layer (BLL) model~\cite{harrison2024variational}, where a deterministic LLM-based feature extractor is followed by random last layer weights for uncertainty representation. Specifically, let $\boldsymbol{\phi}_{\small\mathbf{W}}(\mathbf{x})\in \mathbb{R}^d$ be the $d$-dimensional feature mapping with ${\mathbf{W}}$ collecting the weights of the LLM. Given training dataset $\mathcal{D}:=\{({\bf x}_n, {\bf y}_n)\}_{n=1}^{N}$ with ${\bf y}_n :=[y_{n,1},\ldots, y_{n,C}]^\top\in \{ 0,1\}^{C\times 1}$ being a one-hot encoding of a $C$-class classification task, the BLL model per sample $n$ is given by
\begin{equation}
p({\bf y}_n|\mathbf{x}_n, \boldsymbol{\Theta}) =  \frac{\exp({\bf y}_n^\top{\bf z}_n)}{{\bf 1}_C^\top \exp({\bf z}_n)}, \ \ \mathbf{z}_n = \boldsymbol{\Theta}\boldsymbol{\phi}_{\mathbf{W}}(\mathbf{x}_n) \label{eq:logits}
\end{equation}
where ${\bf 1}_C$ is a $C\times 1$ all-one vector, $\boldsymbol{\Theta} = [\boldsymbol{\theta}_1, \ldots, \boldsymbol{\theta}_C]^\top \in \mathbb{R}^{C \times d}$ is the classification weight matrix with $\boldsymbol{\theta}_c \in \mathbb{R}^d$ being the {\it random} weight vector for class $c$ with iid Gaussian prior 
\begin{align}
  p(\boldsymbol{\Theta}) = \prod_{c=1}^C \mathcal{N}(\boldsymbol{\theta}_c; \mathbf{0}, \sigma_\theta^2 \mathbf{I}_d)  \label{eq:Theta_prior}
\end{align}
where the prior variance $\sigma_\theta^2$ is a {\it hyperparameter} to be tuned.



Direct optimization of the marginal likelihood $p(\mathcal{D}) = \int p(\mathcal{D}|\boldsymbol{\Theta}) p(\boldsymbol{\Theta}) d\boldsymbol{\Theta}$ is intractable due to the nonlinear softmax-based likelihood function in~\eqref{eq:logits}. Moreover, gradient computation would require the full marginal likelihood, making mini-batch training impossible, and the flexibility of neural network features can lead to over-concentration of the posterior. To address these issues, we rely on the  variational inference framework that jointly seeks the model parameters ${\bf W}$ and parameter posterior $q(\boldsymbol{\Theta})$ by maximizing the evidence lower bound (ELBO)
\begin{align}
\mathcal{L}^\text{ELBO}(q(\boldsymbol{\Theta}),{\bf W};\mathcal{D} ) = &\mathbb{E}_{q(\boldsymbol{\Theta})} [\log p(\mathcal{D}|\boldsymbol{\Theta})] \nonumber \\
&- \text{KL}(q(\boldsymbol{\Theta}) \| p(\boldsymbol{\Theta})) 
\label{eq:elbo_general}
\end{align}

where 
\begin{align}
 \log p(\mathcal{D}|\boldsymbol{\Theta}) 
 &= \log \prod_{n=1}^N \log p({\bf y}_n|\mathbf{x}_n, \boldsymbol{\Theta}) \nonumber \\
 &= \sum_{n = 1}^N {\bf y}_n^\top{\bf z}_n - \sum_{n=1}^N \log ({\bf 1}_C^\top \exp ({\bf z}_n))\label{eq:batch likelihood}
\end{align}

For the sake of tractability, the approximate posterior of $\boldsymbol{\Theta}$ will be assumed to be factorizable across classes and the per-class parameter posterior will be approximated by a Gaussian with mean $\boldsymbol{\mu}_c$ and covariance $\mathbf{S}_c$, namely,
\begin{equation}
q(\boldsymbol{\Theta}) = \prod_{c=1}^C q(\boldsymbol{\Theta}_c) = \prod_{c=1}^C \mathcal{N}(\boldsymbol{\theta}_c; \boldsymbol{\mu}_c, \mathbf{S}_c) \label{eq:variational_posterior}
\end{equation}
where we assume factorization across classes (reducing computational complexity) while retaining full covariance $\mathbf{S}_c \in \mathbb{R}^{d \times d}$ within each class (capturing feature correlations).

Taking the expectation of (\ref{eq:batch likelihood}) wrt $q(\boldsymbol{\Theta})$ in (\ref{eq:variational_posterior}) is intractable due to the log-softmax. We will apply again Jensen's inequality to yield a lower bound as
\begin{align}
&\mathbb{E}_{q(\boldsymbol{\Theta})} \left[ - \log \sum_{c=1}^C \exp(z_{n,c})\right] 
{\geq}  \nonumber \\
&- \log \mathbb{E}\left[\sum_{c=1}^C \exp(z_{n,c})\right] 
=  - \log \sum_{c=1}^C \mathbb{E}\left[\exp(z_{n,c})\right]\;\label{eq:jensen_bound}
\end{align}
Since $z_{n,c} = \boldsymbol{\theta}_c^\top \boldsymbol{\phi}_{\mathbf{W}}(\mathbf{x}_n)$ and $\boldsymbol{\theta}_c \sim \mathcal{N}(\boldsymbol{\mu}_c, \mathbf{S}_c)$, we have
\begin{align}
\mathbb{E}[z_{n,c}] &= \boldsymbol{\mu}_c^\top \boldsymbol{\phi}_{\mathbf{W}}(\mathbf{x}_n) \label{eq:mean_logit}\\
\mathbb{E}[\exp(z_{n,c})] &=\label{eq:mgf}\\
&\!\!\!\!\exp\left(\boldsymbol{\mu}_c^\top \boldsymbol{\phi}_{\mathbf{W}}(\mathbf{x}_n) 
+ \frac{1}{2}\boldsymbol{\phi}_{\mathbf{W}}(\mathbf{x}_n)^\top \mathbf{S}_c \boldsymbol{\phi}_{\mathbf{W}}(\mathbf{x}_n)\right) \nonumber
\end{align}

Further, the KL divergence term is expressed explicitly as
\begin{align}
\text{KL}(q(\boldsymbol{\Theta}) \| &p(\boldsymbol{\Theta})) = \sum_{c=1}^C \text{KL}(\mathcal{N}(\boldsymbol{\mu}_c, \mathbf{S}_c) \| \mathcal{N}(\mathbf{0}, \sigma_\theta^2 \mathbf{I}_d)) \nonumber\\
& = \sum_{c = 1}^C\left(\frac{1}{2\sigma_\theta^2}\left(\text{tr}(\mathbf{S}_c) + \boldsymbol{\mu}_c^\top\boldsymbol{\mu}_c\right) - \frac{1}{2}\log|\mathbf{S}_c| \right) \nonumber\\
&\quad + \frac{dC}{2}\log\sigma_\theta^2 - \frac{dC}{2} \label{eq:kl_closed}
\end{align}

Combining all terms, the ELBO objective is given by
\begin{align}
&\mathcal{L}^\text{ELBO}(\boldsymbol{\Psi}, {\bf W}; \mathcal{D})\label{eq:elbo_final} \\
&= \sum_{n=1}^{N} \Bigg[ \sum_{c=1}^C y_{n,c} \boldsymbol{\mu}_c^\top \boldsymbol{\phi}_{\mathbf{W}}(\mathbf{x}_n) \nonumber\\
&\qquad - \text{LSE}_c\Big(\boldsymbol{\mu}_c^\top \boldsymbol{\phi}_{\mathbf{W}}(\mathbf{x}_n) + \tfrac{1}{2}\boldsymbol{\phi}_{\mathbf{W}}(\mathbf{x}_n)^\top \mathbf{S}_c \boldsymbol{\phi}_{\mathbf{W}}(\mathbf{x}_n)\Big) \Bigg]\nonumber\\
&\quad - \sum_{c=1}^C \left[ \frac{1}{2\sigma_\theta^2}\left(\text{tr}(\mathbf{S}_c) + \boldsymbol{\mu}_c^\top\boldsymbol{\mu}_c\right) - \frac{1}{2}\log|\mathbf{S}_c| \right] + \text{const} \nonumber
\end{align}
where $\boldsymbol{\Psi}:=\{(\boldsymbol{\mu}_c, \mathbf{S}_c)\}_{c=1}^C$ collects the variational parameters and $\text{LSE}_c(\cdot) = \log \sum_{c=1}^C \exp(\cdot)$ denotes the log-sum-exp function with sum over class $c$, which provides numerical stability when computing the logarithm of sums of exponentials. 

Given a pre-trained LLM with weights ${\bf W}_0$, the fine-tuned weight parameterization is typically given by ${\bf W}:= {\bf W}_0 + \Delta {\bf W}$. For PEFT, $\Delta {\bf W}$ is typically sought as a low-rank representation. 
However, standard LoRA-based approaches suffer from geometric limitations that undermine uncertainty estimation in the BLL model, as we discuss next.

\subsection{Fine-Tuned LLM via Orthogonalized LoRA}


Recent work on UQ has established that BLL methods critically depend on the geometric properties of the learned features. In particular, SNGP~\citep{liu2020simple} demonstrates that \emph{distance-aware features}, where semantically distinct inputs remain well-separated in feature space, are essential for reliable uncertainty estimation. When the feature extractor suffers from \emph{feature collapse}~\citep{postels2021practicality}, projecting diverse inputs onto a narrow subspace, the BLL model cannot distinguish in-distribution from out-of-distribution samples, as the necessary distance information is lost during feature extraction.

Standard LoRA however, struggles to maintain this property due to its well-documented tendency toward rank collapse~\citep{zhang2025polar}, where the stable rank approaches 1.0 despite nominally higher allocated ranks; see Figure~\ref{fig:stable_rank_analysis}. This geometric compression directly violates the distance-awareness requirement, yielding the suboptimal UQ performance of existing LoRA-based Bayesian methods; see Figure~\ref{fig:ablation study}(a). To alleviate this issue, we adapt the recently developed orthogonalized low-rank adapters that preserve multi-directional feature geometry.

\textbf{Orthogonal Parametrization.}
We leverage the PoLAR parameterization, where the additive weight for a particular layer $\Delta \mathbf{W} \in \mathbb{R}^{m \times n} $ is given by
\begin{equation}
\Delta \mathbf{W} = \mathbf{U}\boldsymbol{\Lambda} \mathbf{V}^\top. \label{eq:Polar}
\end{equation}
Here, $\mathbf{U} \in \text{St}(m, r)$, $\mathbf{V} \in \text{St}(n, r)$, and $\bfLambda \in \mathbb{R}^{r \times r}$ is unconstrained for effective optimization with $\text{St}(m, r):= \{\mathbf{M} \in \mathbb{R}^{m \times r} | \mathbf{M}^\top\mathbf{M}  = \I_r \}$ denoting a Stiefel manifold, i.e., matrices with orthonormal columns. These orthogonality constraints effectively prevent rank collapse when optimized properly.
Integrating PoLAR into the VBLL framework, we will adapt the ELBO objective (\ref{eq:elbo_final}) by setting ${\bf W}:= {\bf W}_0 + \Delta \W$ with $\Delta \W$ parametrized by PoLAR~\eqref{eq:Polar}. Thus, the resulting PoLAR-VBLL jointly seek the PoLAR parameters $\boldsymbol{\Psi}_{\text{polar}} := \{\mathbf{U}, \boldsymbol{\Lambda}, \mathbf{V}\}$ and the variational parameters $\boldsymbol{\Psi}$ via
\begin{align}
\{\hat{\boldsymbol{\Psi}}, \hat{\boldsymbol{\Psi}}_{\text{polar}}\} 
&= \underset{\boldsymbol{\Psi},\boldsymbol{\Psi}_{\text{polar}}}{\arg\max}\ \ \mathcal{L}^\text{ELBO}(\boldsymbol{\Psi},\boldsymbol{\Psi}_{\text{polar}}; \mathcal{D}) \nonumber \\
&\quad {\rm s.to}\  \mathbf{U} \in \text{St}(m, r), \mathbf{V} \in \text{St}(n, r)  \;\label{eq:elbo_final_PoLAR}
\end{align}

\textbf{Scalable Optimization via Landing Fields.} To cope with the manifold constraints on $\U$ and $\V$, standard approaches rely on Riemannian optimization, which involves retraction operations. On Stiefel manifolds, these retractions require either Singular Value Decomposition (SVD) or QR decomposition, making them impractical for large-scale models. This computational bottleneck can be alleviated using landing methods \citep{gao2022optimization,schechtman2023orthogonal}. For instance, optimizing $\mathbf{U}$ simply requires to replace its Euclidean gradient with the so-termed landing field:
\begin{equation}\label{eq.landing-field}
\boldsymbol{\Gamma}(\mathbf{U}) = \boldsymbol{\psi}(\mathbf{U})\mathbf{U} + \lambda \nabla N(\mathbf{U})
\end{equation}
where $\boldsymbol{\psi}(\mathbf{U}) = \text{Skew}(\nabla_{\mathbf{U}} \mathcal{L}(\mathbf{U}, \boldsymbol{\Lambda}, \mathbf{V})\mathbf{U}^\top)$ is the (generalized) Riemannian gradient component and $\nabla N(\mathbf{U}) = 4\mathbf{U}(\mathbf{U}^\top \mathbf{U} - \mathbf{I}_r)$ is the gradient of the infeasibility penalty $N(\mathbf{U}) = \|\mathbf{U}^\top \mathbf{U} - \mathbf{I}_r\|_F^2$. The parameter $\lambda > 0$ controls the strength of penalization for constraint violations. In other words, landing is an infeasible method, but with a properly chosen $\lambda$, the constraints are satisfied asymptotically at convergence. By avoiding costly SVD operations, this approach achieves a 3$\times$ to 18$\times$ speedup compared to retraction-based methods on GPUs, depending on the chosen rank.

The combination of orthogonal parameterization and scalable optimization yields theoretical benefits. Notably, PoLAR has been shown, under some assumptions, to converge faster as the rank $r$ increases, in stark contrast to LoRA \citep{zhang2025polar}. This improved scaling with $r$ enables the design of more expressive feature extractors tailored to available memory budgets, thereby justifying our adoption of PoLAR.

\textbf{Joint Optimization of PoLAR-VBLL.} To solve the optimization problem in (\ref{eq:elbo_final_PoLAR}), we will adopt alternating optimization, that consists of the following two steps per iteration.
\begin{itemize}
    \item \textit{Variational Posterior Update:} The gradients with respect to the variational parameters $\boldsymbol{\Psi}_c$ follow standard variational training procedures (see Eqs.~(\ref{eq:grad_mu})-(\ref{eq:grad_S}) in App.~\ref{sec:detailed_gradients});
    \item \textit{PoLAR Parameter Update:} For the PoLAR parameters constrained to Stiefel manifolds, we employ landing field (cf. Eq. \ref{eq.landing-field}) to avoid expensive retraction operations.
   See App.~\ref{sec:detailed_gradients} (Eqs.~(\ref{eq:grad_Lambda})–(\ref{eq:landing_V})) for detailed derivations of the Riemannian gradients and updates.    
\end{itemize}

The unified framework, together with infeasible Riemannian optimization for computational efficiency, yields a feature extractor that enhances both downstream performance and the reliability of UQ. Please refer to Algorithms~\ref{alg:polar_vbll_training} and \ref{alg:polar_vbll_inference} in the Appendix for the detailed implementations. 



\subsection{Uncertainty-aware predictive inference}
Having available the parameter estimates after PoLAR-VBLL training, we are ready to predict for the label ${ y}\in\{1,\ldots,C \}$ for any given test input ${\bf x}$. Specifically, this predictive pdf is given by
\begin{equation}
p({ y}|\mathbf{x}, \mathcal{D}) = \int_{\boldsymbol{\Theta}} p({ y}|\boldsymbol{\Theta}, {\bf x}) q(\boldsymbol{\Theta}) d\boldsymbol{\Theta} \approx \frac{1}{K} \sum_{k=1}^K p({ y}|\mathbf{x}, \boldsymbol{\Theta}^{(k)}) \label{eq:predictive}
\end{equation}
where we have employed Monte Carlo sampling to approximate the integral via $\boldsymbol{\Theta}^{(k)} \sim q(\boldsymbol{\Theta})$ and 
$p(y|\mathbf{x}, \boldsymbol{\Theta}^{(k)}) = \text{softmax}(\boldsymbol{\Theta}^{(k)}\boldsymbol{\phi}_{\hat{\mathbf{W}}}(\mathbf{x}))$
with $\hat{\mathbf{W}} = \mathbf{W}_0 + \hat{\mathbf{U}}\hat{\boldsymbol{\Lambda}}\hat{\mathbf{V}}^\top$.

While the PoLAR-VBLL framework provides an efficient method for end-to-end training, and our ablation confirms that the Jensen-tightened ELBO remains tight throughout training (Appendix Table~\ref{tab:jensen_vs_mc}), the variational covariance $\hat{\mathbf{S}}_c$ is learned through gradient-based optimization, which captures global trends across the training trajectory but may not fully reflect the precise local geometry at the converged mode. To complement this learned covariance with exact curvature information, we optionally introduce a post-hoc LA step that computes the Hessian of the log-posterior at the VBLL-discovered mode $\hat{\boldsymbol{\mu}}_c$.
Specifically, given the estimated PoLAR parameters $\hat{\boldsymbol{\Psi}}_{\text{polar}}$, we will evaluate the Hessian of $\log p({\cal D}, \boldsymbol{\Theta}|\hat{\boldsymbol{\Psi}}_{\text{polar}})$ at the posterior mean $\{\hat{\boldsymbol{\mu}}_c\}_c$ as
\begin{equation}
    \mathbf{H} = -\nabla^2_{\boldsymbol{\Theta}} \left( \log p(\mathcal{D} | \boldsymbol{\Theta}, \hat{\boldsymbol{\Psi}}_{\text{polar}})+\log p(\boldsymbol{\Theta}) \right)\Big\rvert_{\boldsymbol{\Theta} = \{\boldsymbol{\mu}_c\}_c}\label{eq:hessian}
\end{equation}
where $\log p(\mathcal{D} | \boldsymbol{\Theta}, \hat{\boldsymbol{\Psi}}_{\text{polar}})$ and $\log p(\boldsymbol{\Theta})$ are given by (\ref{eq:batch likelihood}) and (\ref{eq:Theta_prior}). 
Note that ${\bf H}$ is also the Bayesian Fisher information matrix of $\boldsymbol{\Theta}$, whose inverse $\boldsymbol{\Sigma} =\mathbf{H^{-1}} $, the Bayesian Cramer-Rao lower bound, can be taken as a covariance matrix for $\boldsymbol{\Theta}$. For the sake of tractability, we will still enforce a factorizable posterior over $\boldsymbol{\Theta}$, by ignoring the off-diagonal elements in $\boldsymbol{\Sigma}$. With $\boldsymbol{\Sigma}_c$ being the matrix on the diagonal of $\boldsymbol{\Sigma}$ corresponding to $\boldsymbol{\theta}_c$, the resulting corrected posterior is
\begin{equation}
    \tilde{q}(\boldsymbol{\theta}_c | \mathcal{D}) = \mathcal{N}(\boldsymbol{\theta}_c; \hat{\boldsymbol{\mu}}_c, \boldsymbol{\Sigma}_c) \label{eq:laplace_posterior}
\end{equation}
which will be used to make the prediction in (\ref{eq:predictive}).


\noindent{\bf Remark.}  Our strategy uses the scalable VBLL framework to first identify a high-quality mode $\hat{\boldsymbol{\mu}}_c$ along with the PoLAR parameters, and then applies LA as a `finishing touch' to better characterize the posterior covariance around this well-chosen point. Notably, the post-hoc LA calibration does not affect the accuracy of the calibrated model~\citep{yang2023bayesian}. This hybrid approach nicely combines the strengths of variational training and post-hoc LA for enhanced uncertainty assessment. We have empirically validated the benefits of this additional step in our ablation studies, demonstrating improved performance on key UQ metrics such as calibration and out-of-distribution detection.



\section{Experimental Results}

In this section, we compare our PoLAR-VBLL with existing methods on real-world datasets. We first introduce the experimental settings, including baselines, fine-tuning protocols, and evaluation procedures. We then evaluate PoLAR-VBLL's uncertainty estimation and generalization abilities in both in-distribution and out-of-distribution scenarios.

\subsection{Settings}

\textbf{Fine-Tuning and Evaluation.} We implement PoLAR-VBLL using the PEFT library~\citep{mangrulkar2022peft} and fine-tune \texttt{LlaMA-3.1-8B}~\citep{touvron2023llama} on common-sense reasoning tasks. Additional results on \texttt{LlaMA2-7B} are provided in Table~\ref{tab:main_results_full}. Following Laplace-LoRA~\citep{yang2023bayesian} and BLoB~\citep{wang2024blob}, we apply PoLAR adapters~\citep{zhang2025polar} to the output layer and the queries and values of all attention layers, with rank $r=8$ for all methods. We adopt default hyperparameters from the PEFT library and the original PoLAR implementation; see Appendix~\ref{sec:Implementation Details} for details. 

We evaluate six common-sense reasoning in-distribution (ID) datasets, namely, Winogrande-Small and Medium (WG-S, WG-M) \citep{sakaguchi2021winogrande}, ARC-Challenge (ARC-C) \citep{clark2018think}, ARC-Easy (ARC-E) \citep{clark2018think}, OpenBookQA (OBQA) \citep{OpenBookQA2018}, BoolQ~\citep{clark2019boolq} and additional chemistry (Chem) and physics (Phy) from the MMLU benchmark~\citep{hendrycks2021ethics,hendryckstest2021} for out-of-distribution (OOD) evaluation. We cast the common-sense reasoning tasks as classification over possible answers and fine-tune the LLM to maximize the ELBO in Eq.~(\ref{eq:elbo_final}). For evaluation, we report Accuracy (ACC), Expected Calibration Error (ECE)~\citep{naeini2015obtaining}, and Negative Log-Likelihood (NLL).


\textbf{Baselines and Implementation Details.} We compare PoLAR-VBLL with a comprehensive set of baselines spanning standard PEFT methods and state-of-the-art uncertainty quantification (UQ) approaches. For standard PEFT baselines, we include Maximum Likelihood Estimation (MLE)~\citep{hu2022lora,myung2003tutorial,le1990maximum} and Maximum A Posteriori (MAP)~\citep{greig1989exact}. For UQ methods applied on top of LoRA fine-tuning, we consider Monte-Carlo Dropout (MCD)~\citep{gal2016dropout}, Deep Ensemble(ENS)~\citep{lakshminarayanan2017simple}, and Laplace-LoRA (LA)~\citep{yang2023bayesian,kristiadi2024sober}. We also compare against recent Bayesian LoRA methods, including BLoB~\citep{wang2024blob}, ScalaBL~\citep{samplawski2025scalable}, C-LoRA~\citep{rahmati2025c}, and TFB~\citep{shi2024training}. For TFB, we additionally report results with last-layer-only uncertainty (TFB-LL). To isolate the contribution of our polar decomposition, we include {\it PoLAR-ized} variants (PoLAR-MLE, PoLAR-LA, PoLAR-LA-LL, PoLAR-BLoB), where we replace standard LoRA with PoLAR decomposition while keeping other components unchanged; implementation details are provided in Appendix~\ref{sec:Implementation Details}. 
We strictly followed the official implementation and hyperparameter configurations for ScalaBL. However, despite our best efforts, our reproduction yielded a model where the low ECE comes at the cost of substantially reduced accuracy, suggesting potential underfitting. Thus, we moved the results of ScalaBL to Table~\ref{tab:llama3_results_full} in the Appendix.
We also attempted PoLAR-TFB, which applies TFB~\citep{shi2024training} to the core matrix $\boldsymbol{\Lambda}$, but found that TFB does not transfer well to PoLAR's compact core matrix.

\subsection{Results on ID and OOD Datasets}

\begin{table*}[t]
\centering
\caption{Performance of different methods on \texttt{Llama-3.1-8B}. ACC and ECE are reported in percentages. The evaluation is done across six in-distribution common-sense reasoning datasets with fine-tuning of 5000 steps. For OOD evaluation, models are trained on OBQA and tested on other datasets. \textbf{Bold} and \underline{underlined} denote the best and second-best performance, respectively.}
\label{tab:main_results_BLOB_10_Only}

\setlength{\tabcolsep}{2.5pt}
\resizebox{\textwidth}{!}{%
\scriptsize
\begin{tabular}{ll|cccccc|cc|cc}
\toprule
& & \multicolumn{6}{c|}{\textbf{In-Distribution Datasets}} & \multicolumn{4}{c}{\textbf{Out-of-Distribution Datasets} (OBQA$\rightarrow$X)} \\
\cmidrule{3-8} \cmidrule{9-12}
& & & & & & & & \multicolumn{2}{c|}{\textit{Small Shift}} & \multicolumn{2}{c}{\textit{Large Shift}} \\
\textbf{Metric} & \textbf{Method} & \textbf{WG-S} & \textbf{ARC-C} & \textbf{ARC-E} & \textbf{WG-M} & \textbf{OBQA} & \textbf{BoolQ} & \textbf{ARC-C} & \textbf{ARC-E} & \textbf{Chem} & \textbf{Phy} \\
\midrule
\multirow{13}{*}{\rotatebox{90}{\textbf{ACC ($\uparrow$)}}}
& MLE & \underline{77.92$_{\pm0.62}$} & 81.05$_{\pm1.62}$ & 90.66$_{\pm0.10}$ & 82.80$_{\pm0.96}$ & \underline{88.30$_{\pm0.36}$} & 87.86$_{\pm0.50}$ & 79.33$_{\pm0.64}$ & 85.66$_{\pm0.50}$ & 48.00$_{\pm2.00}$ & 43.33$_{\pm1.53}$ \\
& LA & \underline{77.92$_{\pm0.62}$} & 81.05$_{\pm1.62}$ & 90.66$_{\pm0.10}$ & 82.80$_{\pm0.96}$ & \underline{88.30$_{\pm0.36}$} & 87.86$_{\pm0.50}$ & 79.33$_{\pm0.64}$ & 85.66$_{\pm0.50}$ & 48.00$_{\pm2.00}$ & 43.33$_{\pm1.53}$ \\
& PoLAR-MLE & \textbf{78.09$_{\pm0.39}$} & \underline{81.21$_{\pm1.02}$} & 90.25$_{\pm1.24}$ & \textbf{83.61$_{\pm0.32}$} & 86.67$_{\pm1.22}$ & \textbf{89.16$_{\pm0.27}$} & \underline{80.97$_{\pm0.76}$} & 85.57$_{\pm0.21}$ & \underline{48.33$_{\pm0.58}$} & 43.33$_{\pm2.52}$ \\
& PoLAR-LA & \textbf{78.09$_{\pm0.39}$} & \underline{81.21$_{\pm1.02}$} & 90.25$_{\pm1.24}$ & \textbf{83.61$_{\pm0.32}$} & 86.67$_{\pm1.22}$ & \textbf{89.16$_{\pm0.27}$} & \underline{80.97$_{\pm0.76}$} & 85.57$_{\pm0.21}$ & \underline{48.33$_{\pm0.58}$} & 43.33$_{\pm2.52}$ \\
& PoLAR-LA-LL & \textbf{78.09$_{\pm0.39}$} & \underline{81.21$_{\pm1.02}$} & 90.25$_{\pm1.24}$ & \textbf{83.61$_{\pm0.32}$} & 86.67$_{\pm1.22}$ & \textbf{89.16$_{\pm0.27}$} & \underline{80.97$_{\pm0.76}$} & 85.57$_{\pm0.21}$ & \underline{48.33$_{\pm0.58}$} & 43.33$_{\pm2.52}$ \\
& TFB-LL & 76.25$_{\pm0.63}$ & 80.58$_{\pm1.10}$ & 91.03$_{\pm0.72}$ & 82.70$_{\pm0.33}$ & \underline{88.30$_{\pm0.56}$} & 87.53$_{\pm0.62}$ & \underline{80.97$_{\pm1.70}$} & \underline{85.74$_{\pm0.47}$} & 47.33$_{\pm2.08}$ & 45.67$_{\pm0.58}$ \\
& TFB & 73.53$_{\pm0.87}$ & 80.31$_{\pm1.15}$ & \underline{91.20$_{\pm1.40}$} & 80.71$_{\pm0.44}$ & 86.80$_{\pm1.06}$ & 87.83$_{\pm0.72}$ & 80.90$_{\pm1.51}$ & 84.50$_{\pm1.03}$ & 46.87$_{\pm1.04}$ & \underline{48.83$_{\pm1.92}$} \\
& C-LoRA & 77.16$_{\pm0.58}$ & 78.95$_{\pm0.53}$ & 90.40$_{\pm1.10}$ & 82.16$_{\pm0.32}$ & 86.83$_{\pm0.43}$ & 88.26$_{\pm0.92}$ & 80.07$_{\pm1.79}$ & 85.04$_{\pm0.36}$ & 47.67$_{\pm3.06}$ & 41.33$_{\pm2.89}$ \\
& BLoB & 72.36$_{\pm0.96}$ & 79.42$_{\pm1.19}$ & 90.16$_{\pm1.07}$ & 79.32$_{\pm0.95}$ & 87.53$_{\pm1.17}$ & 87.54$_{\pm0.54}$ & 79.10$_{\pm0.91}$ & 84.20$_{\pm1.01}$ & 45.67$_{\pm4.51}$ & 45.67$_{\pm0.58}$ \\
& PoLAR-BLoB & 76.49$_{\pm0.34}$ & 80.03$_{\pm1.59}$ & 91.19$_{\pm0.27}$ & 82.29$_{\pm0.37}$ & 87.67$_{\pm0.46}$ & 87.73$_{\pm0.82}$ & 80.34$_{\pm1.33}$ & 84.29$_{\pm1.08}$ & 46.18$_{\pm3.18}$ & 46.88$_{\pm2.09}$ \\
\cmidrule{2-12}
& PoLAR-VBLL (w/o LA) & 77.26$_{\pm0.50}$ & \textbf{81.79$_{\pm0.42}$} & \textbf{91.38$_{\pm0.39}$} & \underline{83.04$_{\pm0.46}$} & \textbf{88.43$_{\pm0.25}$} & \underline{88.88$_{\pm0.43}$} & \textbf{81.11$_{\pm0.82}$} & \textbf{85.92$_{\pm0.50}$} & \textbf{49.30$_{\pm1.61}$} & \textbf{48.91$_{\pm1.05}$} \\
& \textbf{PoLAR-VBLL} & 77.26$_{\pm0.50}$ & \textbf{81.79$_{\pm0.42}$} & \textbf{91.38$_{\pm0.39}$} & \underline{83.04$_{\pm0.46}$} & \textbf{88.43$_{\pm0.25}$} & \underline{88.88$_{\pm0.43}$} & \textbf{81.11$_{\pm0.82}$} & \textbf{85.92$_{\pm0.50}$} & \textbf{49.30$_{\pm1.61}$} & \textbf{48.91$_{\pm1.05}$} \\
\midrule
\multirow{13}{*}{\rotatebox{90}{\textbf{ECE ($\downarrow$)}}}
& MLE & 21.11$_{\pm0.56}$ & 17.95$_{\pm1.83}$ & 8.95$_{\pm0.21}$ & 15.46$_{\pm0.71}$ & 8.33$_{\pm0.19}$ & 4.76$_{\pm0.60}$ & 13.89$_{\pm1.04}$ & 10.31$_{\pm1.06}$ & 28.73$_{\pm1.02}$ & 37.02$_{\pm2.77}$ \\
& LA & 16.41$_{\pm1.20}$ & 9.72$_{\pm1.28}$ & 4.51$_{\pm0.31}$ & 8.37$_{\pm0.82}$ & 7.40$_{\pm0.13}$ & 2.33$_{\pm0.42}$ & 5.85$_{\pm2.05}$ & 5.09$_{\pm0.17}$ & 11.69$_{\pm0.91}$ & 13.02$_{\pm2.87}$ \\
& PoLAR-MLE & 20.48$_{\pm0.99}$ & 16.99$_{\pm1.77}$ & 9.02$_{\pm0.93}$ & 18.64$_{\pm1.21}$ & 8.56$_{\pm1.50}$ & 2.12$_{\pm0.21}$ & 10.84$_{\pm0.39}$ & 8.35$_{\pm0.63}$ & 26.33$_{\pm2.01}$ & 33.22$_{\pm1.27}$ \\
& PoLAR-LA & 15.19$_{\pm5.43}$ & 9.69$_{\pm1.06}$ & 6.15$_{\pm1.06}$ & \textbf{3.16$_{\pm0.65}$} & 6.04$_{\pm0.41}$ & \underline{1.88$_{\pm0.29}$} & 7.09$_{\pm1.63}$ & 5.19$_{\pm0.52}$ & \underline{11.48$_{\pm1.07}$} & 16.09$_{\pm3.10}$ \\
& PoLAR-LA-LL & 15.06$_{\pm9.29}$ & 8.36$_{\pm1.76}$ & 5.41$_{\pm0.10}$ & 8.30$_{\pm0.79}$ & 6.21$_{\pm0.69}$ & 1.96$_{\pm0.24}$ & 9.45$_{\pm0.83}$ & 7.27$_{\pm0.97}$ & 14.89$_{\pm1.12}$ & 13.75$_{\pm1.84}$ \\
& TFB-LL & 12.34$_{\pm0.36}$ & 10.43$_{\pm1.69}$ & 3.77$_{\pm0.12}$ & 8.02$_{\pm0.86}$ & 4.36$_{\pm0.53}$ & 3.13$_{\pm0.71}$ & 7.86$_{\pm1.69}$ & 5.35$_{\pm0.81}$ & 18.75$_{\pm4.23}$ & 20.67$_{\pm3.02}$ \\
& TFB & 5.90$_{\pm0.56}$ & \underline{4.96$_{\pm1.45}$} & \underline{3.72$_{\pm0.22}$} & \underline{3.28$_{\pm0.64}$} & 6.18$_{\pm0.43}$ & 4.21$_{\pm0.42}$ & 5.10$_{\pm0.37}$ & 4.03$_{\pm1.25}$ & 17.83$_{\pm2.71}$ & 15.80$_{\pm2.32}$ \\
& C-LoRA & 18.31$_{\pm0.42}$ & 8.13$_{\pm0.79}$ & 5.42$_{\pm0.16}$ & 6.22$_{\pm1.07}$ & 5.33$_{\pm0.75}$ & 3.63$_{\pm0.63}$ & 13.94$_{\pm1.99}$ & 10.38$_{\pm0.87}$ & 27.99$_{\pm4.78}$ & 35.25$_{\pm3.44}$ \\
& BLoB & 5.78$_{\pm0.75}$ & 7.34$_{\pm1.31}$ & 5.66$_{\pm0.65}$ & 3.91$_{\pm0.93}$ & 5.30$_{\pm1.72}$ & 2.61$_{\pm0.49}$ & \underline{4.87$_{\pm0.56}$} & 5.05$_{\pm0.57}$ & 13.23$_{\pm3.50}$ & \underline{11.31$_{\pm1.86}$} \\
& PoLAR-BLoB & \underline{5.21$_{\pm0.92}$} & 7.02$_{\pm0.96}$ & 5.76$_{\pm0.33}$ & 7.70$_{\pm1.07}$ & \underline{2.36$_{\pm0.69}$} & 2.38$_{\pm0.36}$ & 6.52$_{\pm0.67}$ & \textbf{3.76$_{\pm0.97}$} & 20.29$_{\pm3.49}$ & 18.43$_{\pm2.32}$ \\
\cmidrule{2-12}
& PoLAR-VBLL (w/o LA) & 8.19$_{\pm1.88}$ & 5.86$_{\pm0.49}$ & 4.90$_{\pm1.43}$ & 8.79$_{\pm0.66}$ & 2.96$_{\pm0.13}$ & 2.14$_{\pm0.19}$ & 7.65$_{\pm1.16}$ & 4.09$_{\pm0.94}$ & 15.05$_{\pm1.67}$ & 15.63$_{\pm1.90}$ \\
& \textbf{PoLAR-VBLL} & \textbf{3.89$_{\pm1.27}$} & \textbf{4.92$_{\pm0.49}$} & \textbf{3.71$_{\pm0.29}$} & 3.77$_{\pm0.53}$ & \textbf{2.34$_{\pm0.66}$} & \textbf{1.77$_{\pm0.50}$} & \textbf{4.55$_{\pm0.22}$} & \underline{3.89$_{\pm0.68}$} & \textbf{10.30$_{\pm1.36}$} & \textbf{11.12$_{\pm1.40}$} \\
\midrule
\multirow{13}{*}{\rotatebox{90}{\textbf{NLL ($\downarrow$)}}}
& MLE & 2.23$_{\pm0.01}$ & 1.75$_{\pm0.09}$ & 0.74$_{\pm0.06}$ & 1.05$_{\pm0.12}$ & 0.49$_{\pm0.01}$ & \underline{0.27$_{\pm0.01}$} & 0.90$_{\pm0.04}$ & 0.63$_{\pm0.06}$ & 1.75$_{\pm0.06}$ & 1.94$_{\pm0.08}$ \\
& LA & \textbf{0.57$_{\pm0.01}$} & 1.04$_{\pm0.02}$ & 0.61$_{\pm0.07}$ & 0.56$_{\pm0.06}$ & 0.39$_{\pm0.01}$ & \underline{0.27$_{\pm0.01}$} & \underline{0.54$_{\pm0.01}$} & \textbf{0.39$_{\pm0.02}$} & 1.19$_{\pm0.02}$ & 1.22$_{\pm0.03}$ \\
& PoLAR-MLE & 1.97$_{\pm0.06}$ & 1.02$_{\pm0.02}$ & 0.79$_{\pm0.03}$ & 0.90$_{\pm0.04}$ & 0.48$_{\pm0.05}$ & \underline{0.27$_{\pm0.00}$} & 0.68$_{\pm0.04}$ & 0.50$_{\pm0.04}$ & 1.56$_{\pm0.06}$ & 1.61$_{\pm0.04}$ \\
& PoLAR-LA & 0.61$_{\pm0.03}$ & 0.73$_{\pm0.07}$ & 0.39$_{\pm0.04}$ & \textbf{0.53$_{\pm0.01}$} & \underline{0.37$_{\pm0.07}$} & \underline{0.27$_{\pm0.01}$} & 0.56$_{\pm0.02}$ & 0.42$_{\pm0.03}$ & \underline{1.18$_{\pm0.04}$} & \textbf{1.20$_{\pm0.04}$} \\
& PoLAR-LA-LL & 0.76$_{\pm0.28}$ & 0.61$_{\pm0.04}$ & 0.36$_{\pm0.02}$ & \underline{0.55$_{\pm0.04}$} & 0.39$_{\pm0.03}$ & \underline{0.27$_{\pm0.00}$} & 0.57$_{\pm0.02}$ & 0.48$_{\pm0.03}$ & 1.42$_{\pm0.03}$ & 1.46$_{\pm0.05}$ \\
& TFB-LL & 0.59$_{\pm0.01}$ & 0.64$_{\pm0.04}$ & 0.35$_{\pm0.06}$ & 0.61$_{\pm0.01}$ & \textbf{0.35$_{\pm0.06}$} & \underline{0.27$_{\pm0.01}$} & 0.58$_{\pm0.03}$ & \textbf{0.39$_{\pm0.01}$} & 1.27$_{\pm0.06}$ & 1.37$_{\pm0.09}$ \\
& TFB & 0.59$_{\pm0.01}$ & \textbf{0.58$_{\pm0.07}$} & \underline{0.33$_{\pm0.03}$} & \underline{0.55$_{\pm0.03}$} & 0.39$_{\pm0.01}$ & \underline{0.27$_{\pm0.01}$} & \underline{0.54$_{\pm0.01}$} & 0.44$_{\pm0.11}$ & 1.23$_{\pm0.03}$ & 1.31$_{\pm0.04}$ \\
& C-LoRA & 0.85$_{\pm0.02}$ & 0.90$_{\pm0.07}$ & 0.35$_{\pm0.01}$ & 0.62$_{\pm0.05}$ & 0.58$_{\pm0.06}$ & 0.29$_{\pm0.02}$ & 0.83$_{\pm0.08}$ & 0.63$_{\pm0.04}$ & 1.69$_{\pm0.07}$ & 1.84$_{\pm0.10}$ \\
& BLoB & \underline{0.58$_{\pm0.01}$} & \underline{0.59$_{\pm0.03}$} & \textbf{0.30$_{\pm0.08}$} & 0.60$_{\pm0.05}$ & \underline{0.37$_{\pm0.01}$} & 0.31$_{\pm0.03}$ & 0.55$_{\pm0.02}$ & \underline{0.41$_{\pm0.01}$} & \underline{1.18$_{\pm0.03}$} & 1.31$_{\pm0.06}$ \\
& PoLAR-BLoB & 0.67$_{\pm0.06}$ & \underline{0.59$_{\pm0.04}$} & 0.35$_{\pm0.01}$ & 0.61$_{\pm0.04}$ & \textbf{0.35$_{\pm0.01}$} & 0.29$_{\pm0.01}$ & 0.55$_{\pm0.02}$ & \underline{0.41$_{\pm0.04}$} & 1.26$_{\pm0.05}$ & \underline{1.21$_{\pm0.03}$} \\
\cmidrule{2-12}
& PoLAR-VBLL (w/o LA) & 0.61$_{\pm0.02}$ & 0.64$_{\pm0.02}$ & 0.36$_{\pm0.01}$ & \underline{0.55$_{\pm0.02}$} & 0.41$_{\pm0.02}$ & 0.32$_{\pm0.03}$ & 0.56$_{\pm0.03}$ & 0.44$_{\pm0.01}$ & 1.21$_{\pm0.08}$ & 1.27$_{\pm0.05}$ \\
& \textbf{PoLAR-VBLL} & \underline{0.58$_{\pm0.01}$} & \textbf{0.58$_{\pm0.05}$} & \underline{0.33$_{\pm0.02}$} & \textbf{0.53$_{\pm0.02}$} & \textbf{0.35$_{\pm0.01}$} & \textbf{0.26$_{\pm0.02}$} & \textbf{0.53$_{\pm0.02}$} & 0.42$_{\pm0.02}$ & \textbf{1.16$_{\pm0.01}$} & \textbf{1.20$_{\pm0.05}$} \\
\bottomrule
\end{tabular}%
}
\label{tab:performance_comparison}
\end{table*}

Table~\ref{tab:main_results_BLOB_10_Only} presents results across six ID common-sense reasoning datasets and four OOD datasets with the most advanced baselines. For OOD evaluation, models are trained on OBQA and tested on datasets with varying degrees of distribution shift. A more comprehensive version can be seen at Table~\ref{tab:llama3_results_full}, including all the baselines.

\textbf{ID Performance.} PoLAR-VBLL achieves the best or second-best ACC across all six ID datasets while simultaneously attaining strong calibration (ECE and NLL) in most cases. Notably, our method does not exhibit the accuracy-calibration trade-off commonly observed in sampling-based variational methods~\citep{wang2024blob}, where increasing inference samples $N$ improves uncertainty estimates at the cost of predictive accuracy; see Table~\ref{tab:main_results_full} for detailed comparisons. The simultaneous improvement in ACC, ECE, and NLL mitigates the overconfidence problem inherent in standard MLE fine-tuning.

\textbf{OOD Performance.} We categorize OOD evaluation into two regimes: smaller shifts (ARC-C, ARC-E), which share the multiple-choice science format with OBQA, and larger shifts (Chem, Phy from MMLU), which introduce college-level domain complexity. 
PoLAR-VBLL achieves competitive or best ACC across all OOD settings, with particularly notable gains under larger distribution shifts. Calibration quality is also maintained: PoLAR-VBLL attains strong ECE and NLL across all OOD datasets, suggesting that the learned uncertainty estimates remain reliable even as data distributions diverge from the training regime.

\textbf{Analysis.} The performance gains stem from the synergistic 
design of our framework: PoLAR preserves feature geometry essential for distance-aware uncertainty estimation, VBLL discovers well-calibrated 
posterior modes through joint optimization, and the optional LA provides further refinement. We validate each component's contribution in the ablation studies below.


\subsection{Ablation Studies}
\begin{figure*}
    \centering
    \includegraphics[width=1.0\linewidth]{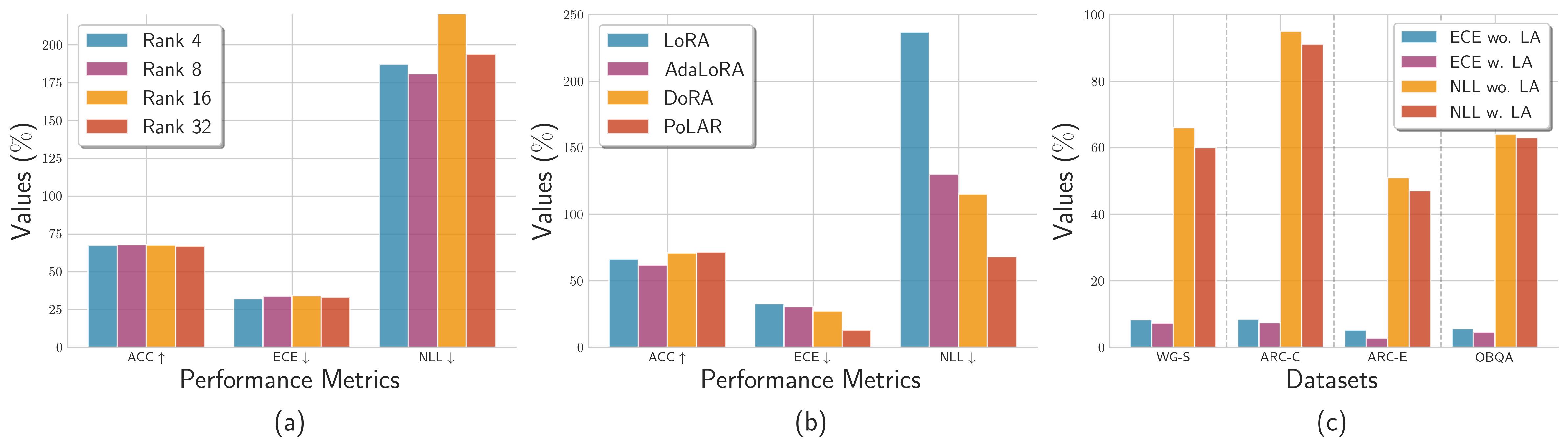}
    \vspace{-5mm}
    \caption{Ablation studies on the WG-S dataset using \texttt{LLaMA 2-7B}: (a) Performance of LoRA-VBLL using different ranks; (b) VBLL coupled with different adapters; and (c) ECE and NLL performances of PoLAR-VBLL with and without LA. }
    \label{fig:ablation study}
\end{figure*}

We conduct ablation studies to validate each component of our PoLAR-VBLL framework. 
The first three experiments are evaluated on the WG-S using the \texttt{LLaMA 2-7B} backbone. 
\paragraph{Rank collapse in standard LoRA}
Our analysis of adapter rank in Figure~\ref{fig:ablation study}(a), standard LoRA demonstrates minimal performance differences across various ranks, with performance remaining relatively flat regardless of rank size, confirming that rank collapse prevents effective utilization of larger rank allocations. This phenomenon is explained by \emph{rank collapse}: Figure~\ref{fig:stable_rank_analysis} (App.~\ref{sec:stable_rank_analysis}) shows that LoRA achieves a stable rank approaching 1, the theoretical minimum, indicating that the learned updates collapse into a nearly rank-1 projection regardless of allocated capacity. In contrast, PoLAR maintains a significantly higher stable rank through orthogonality constraints, ensuring that increased rank translates to genuinely expanded representational capacity rather than redundant directions.


\paragraph{Comparison of adapter architectures.}
Given PoLAR's advantage in preserving rank diversity, we benchmarked it against other PEFT methods, ranging from LoRA to the more advanced adapters such as AdaLoRA~\citep{zhang2023adalora} and DoRA~\citep{liu2024dora}, within the same VBLL framework. As shown in Figure~\ref{fig:ablation study}, PoLAR achieves superior performance across accuracy, ECE, and NLL. These results confirm that PoLAR's more expressive feature representation provides a better foundation not only for the prediction task but also for subsequent uncertainty estimation.

\paragraph{VBLL as the primary driver of calibration.}
To clarify the individual contributions of each component, we refer to the comprehensive comparison in Table~\ref{tab:main_results_BLOB_10_Only}. The results demonstrate that VBLL is the primary driver of UQ: comparing PoLAR-VBLL (w/o LA) with PoLAR-LA-LL, both of which operate on identical architectural scope (last layer only), we find that PoLAR-VBLL (w/o LA) consistently achieves superior calibration across all datasets despite not using any Laplace refinement. 
To further isolate VBLL's contribution from the adapter architecture, we compare different UQ methods under identical PoLAR adapters in Table~\ref{tab:main_results_BLOB_10_Only} and Table~\ref{tab:ablation_blob} (App.~\ref{sec:ablation_blob}) over \texttt{LLaMA-3.1-8B} and \texttt{LLaMA-2-7B}, respectively. PoLAR-VBLL consistently achieves superior ECE compared to other PoLAR-based variants, confirming that the improvements are attributable to the variational training framework rather than solely to the adapter choice.

\paragraph{Efficacy of post-hoc Laplace refinement.}
While VBLL alone achieves strong calibration, the optional LA step provides consistent further improvements. Both Table~\ref{tab:main_results_BLOB_10_Only} and Figure~\ref{fig:ablation study}(c) compare PoLAR-VBLL with and without LA, demonstrating consistent ECE improvements on \texttt{LLaMA-3.1-8B} and \texttt{LLaMA-2-7B}, respectively. Notably, our full PoLAR-VBLL, which applies LA only to the last layer, achieves better calibration than PoLAR-LA applied across all adapter layers. This indicates that post-hoc LA alone cannot match the calibration quality achieved by variational training. VBLL actively guides optimization toward high-quality posterior modes, providing a well-calibrated initialization that enables LA to effectively refine the local covariance structure. Further results in Table~\ref{tab:ablation_blob} in App.~\ref{sec:ablation_blob} confirm that combining VBLL with post-hoc LA yields the best calibration.

\paragraph{Tightness of Jensen bound.}
To validate the tightness of the Jensen bound used in our ELBO formulation, we provide empirical comparison in Table~\ref{tab:jensen_vs_mc} (Appendix~\ref{sec:jensen_bound_tightness}). The comparison between our analytical Jensen-based estimator and a 50-sample Monte Carlo estimator across 400 training steps shows that the absolute gap rapidly converges from 8.27 at initialization to below 0.35 after 50 steps, and remains stable throughout training without any divergence trend. This confirms that the Jensen bound provides a sufficiently tight approximation to the true ELBO objective, enabling efficient closed-form optimization without sacrificing fidelity.
\paragraph{Computational efficiency.}
Finally, we evaluate the runtime of competing methods during inference. Table~\ref{tab:uncertainty_ablation_GPU} (App.~\ref{sec: Memory usage and Run time}) shows that PoLAR-VBLL achieves approximately an order of magnitude speedup compared to BLoB-based methods. This efficiency stems from the LL-only sampling strategy: while BLoB requires multiple complete forward passes through the entire LLM backbone during testing, our approach performs a single backbone pass followed by lightweight sampling of the LL parameters only. For memory use, PoLAR-VBLL maintains a competitive footprint significantly lower than full-network LA. Table~\ref{tab:variational_params} (App.~\ref{sec: Memory usage and Run time}) further details the variational parameter counts, explaining the GPU memory difference between different adapters and VI methods. 

Additional sensitivity analysis on the prior scale $\sigma_\theta$ in Table~\ref{tab:prior_sensitivity} (App.~\ref{app:sensitivity}) confirms that PoLAR-VBLL is robust across different prior configurations.

\section{Conclusions}

This paper introduced PoLAR-VBLL, a scalable and unified framework for uncertainty-aware fine-tuning of LLMs. 
PoLAR addresses the rank collapse issue in conventional adapters through orthogonality constraints, yielding a more expressive feature extractor. 
Building on this foundation, VBLL enables efficient, sampling-free Bayesian training on the final layer for principled UQ. Extensive experiments demonstrate that PoLAR-VBLL consistently outperforms state-of-the-art baselines in both accuracy and uncertainty calibration.
This work presents a principled and practical pathway towards developing more reliable and trustworthy fine-tuned LLMs for real-world applications.
\newpage
\section*{Impact Statement}

This paper aims to improve the reliability and trustworthiness of fine-tuned large language models through principled uncertainty quantification. 
By enabling LLMs to provide well-calibrated confidence estimates, our work has the potential to benefit safety-critical applications such as medical diagnosis assistance, legal document analysis, and autonomous decision-making systems, where understanding when a model is uncertain is crucial for appropriate human oversight.

We do not foresee direct negative societal consequences of this work. 
However, we note that improved uncertainty quantification, while beneficial, should not be viewed as a complete solution to LLM reliability---users should remain aware that even well-calibrated models can still produce incorrect outputs.

\bibliography{example_paper}
\bibliographystyle{icml2026}

\newpage
\appendix
\onecolumn
\section{Computational Complexity Analysis}

We provide a detailed computational complexity analysis for each phase of our PoLAR-VBLL framework, demonstrating its efficiency compared to alternative approaches for uncertainty quantification in fine-tuned LLMs.

\subsection{Training Phase Complexity}

The joint optimization of PoLAR adapter parameters $\boldsymbol{\Psi}_{\text{polar}}: = \{\mathbf{U}, \boldsymbol{\Lambda}, \mathbf{V}\}$ and variational parameters $\boldsymbol{\Psi} = \{\boldsymbol{\mu}_1, \ldots, \boldsymbol{\mu}_C, \mathbf{S}_1, \ldots, \mathbf{S}_C\}$ involves the following computational costs per training iteration:

\noindent{\bf Feature Extraction:} Computing LLM features $\boldsymbol{\phi}_{\mathbf{W}}(\mathbf{x}_n)$ where $\mathbf{W} = \mathbf{W}_0 + \mathbf{U}\boldsymbol{\Lambda}\mathbf{V}^\top$ for a batch of size $B$ requires $\mathcal{O}(B \cdot \text{LLM}_{\text{cost}})$ operations, where $\text{LLM}_{\text{cost}}$ represents the computational cost of a single forward pass through the base language model.

\noindent{\bf PoLAR Parameter Updates:} The landing field optimization on Stiefel manifolds incurs:
\begin{itemize}
    \item Euclidean Gradient Calculation:
    $\mathcal{O}(mnr)$
    \item Riemannian gradient computation: $\mathcal{O}(m^2r + n^2r)$ for skew-symmetric operations $\boldsymbol{\psi}(\mathbf{U})$ and $\boldsymbol{\psi}(\mathbf{V})$
    \item Constraint gradient computation: $\mathcal{O}((m + n)r^{2})$ for infeasibility penalties $\nabla N(\mathbf{U}) = 4\mathbf{U}(\mathbf{U}^\top \mathbf{U} - \mathbf{I}_r)$ and $\nabla N(\mathbf{V})$
    \item Parameter updates: $\mathcal{O}(r(m + n + r))$ for $\mathbf{U}$, $\mathbf{V}$, and $\boldsymbol{\Lambda}$ updates via landing field method
\end{itemize}

\noindent{\bf VBLL Parameter Updates:} Variational inference optimization requires:
\begin{itemize}
    \item ELBO computation: $\mathcal{O}(B \cdot C \cdot d)$ for likelihood terms and log-sum-exp operations in Eq.~\ref{eq:elbo_final_PoLAR}
    \item KL divergence computation: $\mathcal{O}(C \cdot d^2)$ for trace and determinant operations in covariance matrices $\mathbf{S}_c$
    \item Gradient computation: $\mathcal{O}(C \cdot d^2)$ for gradients with respect to variational means $\boldsymbol{\mu}_c$ and covariances $\mathbf{S}_c$
    \item Parameter updates: $\mathcal{O}(B\cdot C \cdot d^2)$ for updating $C$ class-specific posterior distributions
\end{itemize}

The total complexity per training iteration is:
\begin{equation}
\mathcal{O}\left(B \cdot \text{LLM}_{\text{cost}} + r(m^{2} + n^{2}) + B \cdot C \cdot d^2 + r(m + n + r)\right)
\end{equation}

For $T$ training iterations, the overall training complexity becomes:
\begin{equation}
\mathcal{O}\left(T \cdot \left(B \cdot \text{LLM}_{\text{cost}} + r(m^{2} + n^{2}) + B\cdot C \cdot d^2 + r(m + n + r)\right)\right) \label{eq:training_complexity}
\end{equation}

\subsection{Predictive Inference Complexity}

The uncertainty-aware prediction phase involves:

\noindent{\bf Optional Laplace Calibration:} Computing the Hessian of the negative log-likelihood for posterior refinement requires $\mathcal{O}(C \cdot d^2)$ operations using Kronecker-factored approximation (KFAC), which is significantly more efficient than the naive $\mathcal{O}((C \cdot d)^2)$ full Hessian computation.

\noindent{\bf Monte Carlo Sampling:} For $K$ posterior samples from $q(\boldsymbol{\Theta})$:
\begin{itemize}
    \item Parameter sampling: $\mathcal{O}(K \cdot C \cdot d^2)$ for sampling from multivariate Gaussians $\mathcal{N}(\boldsymbol{\mu}_c, \mathbf{S}_c)$
    \item Forward computation: $\mathcal{O}(K \cdot C \cdot d)$ for logit computation $\mathbf{z} = \boldsymbol{\Theta}\boldsymbol{\phi}_{\mathbf{W}}(\mathbf{x}^*)$ and softmax normalization
\end{itemize}

Total inference complexity per test point:
\begin{equation}
\mathcal{O}\left(\text{LLM}_{\text{cost}} + C \cdot d^2 + K \cdot C \cdot d^2\right) \label{eq:inference_complexity}
\end{equation}

\subsection{Comparison with Baseline Methods}

\noindent{\bf vs. Standard LoRA:} Our PoLAR parameterization adds $\mathcal{O}(r^2)$ overhead per update compared to LoRA's $\mathcal{O}(r)$ due to orthogonality constraints, but this is negligible when $r \ll d$ while providing substantially improved stable rank utilization.

\noindent{\bf vs. BLoB:} BLoB requires expensive Monte Carlo sampling during training with $\mathcal{O}(N_{\text{MC}} \cdot \text{LLM}_{\text{cost}})$ cost per ELBO evaluation, where $N_{\text{MC}}$ is the number of Monte Carlo samples. Our VBLL approach achieves analytical ELBO computation, eliminating this sampling overhead.

\noindent{\bf vs. Ensemble Methods:} Maintaining $M$ separate adapter copies requires $\mathcal{O}(M \cdot r(m+n))$ storage and $\mathcal{O}(M \cdot \text{LLM}_{\text{cost}})$ inference time. Our Bayesian approach achieves comparable uncertainty quality with $\mathcal{O}(C \cdot d^2)$ additional parameters.

\noindent{\bf vs. Laplace-LoRA:} Post-hoc Laplace approximation around suboptimal MAP estimates requires similar Hessian computation but lacks the joint optimization benefits of our integrated approach.

\subsection{Memory Complexity}

The space complexity of our framework is:
\begin{equation}
\mathcal{O}\left(|\mathbf{W}_0| + r(m + n + r) + C \cdot d^2 + B \cdot d\right)
\end{equation}

where $|\mathbf{W}_0|$ represents the frozen pre-trained model size, $r(m + n + r)$ accounts for PoLAR parameters, $C \cdot d^2$ stores VBLL covariance matrices, and $B \cdot d$ handles intermediate feature storage during batch processing.

\subsection{Detailed Algorithm Specifications}

\begin{algorithm}[t]
\caption{PoLAR-VBLL Training }
\label{alg:polar_vbll_training}
\begin{algorithmic}[1]
\REQUIRE Pre-trained LLM weights $\mathbf{W}_0$, training dataset $\mathcal{D} = \{(\mathbf{x}_n, {\bf y}_n)\}_{n=1}^{N}$, rank $r$, hyperparameters $\{\eta_{\text{polar}}, \eta_{\text{vbll}}, \lambda, \sigma_\theta^2\}$
\ENSURE Converged PoLAR parameters $\{\hat{\mathbf{U}}, \hat{\boldsymbol{\Lambda}}, \hat{\mathbf{V}}\}$, variational posterior $q(\boldsymbol{\Theta})$
\STATE Initialize $\mathbf{U}_0 \in \text{St}(m,r)$, $\mathbf{V}_0 \in \text{St}(n,r)$ via QR decomposition of random matrices
\STATE Initialize $\boldsymbol{\Lambda}_0 \sim \mathcal{N}(\mathbf{0}, 0.01^2 \mathbf{I}_{r \times r})$
\STATE Initialize VBLL parameters: $\boldsymbol{\mu}_{c,0} = \mathbf{w}_{\text{pretrain},c}$, $\mathbf{S}_{c,0} = \sigma_\theta^2 \mathbf{I}_d$ for $c = 1, \ldots, C$
\FOR{$t = 0, 1, \ldots, T-1$}
    \STATE Sample mini-batch $\mathcal{B}_t \subset \mathcal{D}$ of size $B$
    \STATE Extract features: $\boldsymbol{\phi}_t(\mathbf{x}_n) = \boldsymbol{\phi}_{\mathbf{W}_0 + \mathbf{U}_t\boldsymbol{\Lambda}_t\mathbf{V}_t^\top}(\mathbf{x}_n)$ for all $\mathbf{x}_n \in \mathcal{B}_t$
    \STATE Compute ELBO: $\mathcal{L}_t^{\text{ELBO}}(\boldsymbol{\Psi}_{\text{polar}}, \boldsymbol{\Psi}; \mathcal{B}_t)$ using Eq.~\ref{eq:elbo_final_PoLAR}
    
    \STATE // PoLAR parameter updates via landing field method
    \STATE Compute weight gradient: $\mathbf{G}_t = \frac{\partial \mathcal{L}_t^{\text{ELBO}}}{\partial (\mathbf{U}_t\boldsymbol{\Lambda}_t\mathbf{V}_t^\top)}$
    \STATE Compute factor gradients:
    \STATE \quad $\nabla_{\boldsymbol{\Lambda}} \mathcal{L}_t^{\text{ELBO}} = \mathbf{U}_t^\top \mathbf{G}_t \mathbf{V}_t$
    \STATE \quad $\nabla_{\mathbf{U}} \mathcal{L}_t^{\text{ELBO}} = \mathbf{G}_t \mathbf{V}_t \boldsymbol{\Lambda}_t^\top$
    \STATE \quad $\nabla_{\mathbf{V}} \mathcal{L}_t^{\text{ELBO}} = \mathbf{G}_t^\top \mathbf{U}_t \boldsymbol{\Lambda}_t$
    
    \STATE Compute Riemannian gradients:
    \STATE \quad $\boldsymbol{\psi}(\mathbf{U}_t) = \text{Skew}(\nabla_{\mathbf{U}} \mathcal{L}_t^{\text{ELBO}} \cdot \mathbf{U}_t^\top)$
    \STATE \quad $\boldsymbol{\psi}(\mathbf{V}_t) = \text{Skew}(\nabla_{\mathbf{V}} \mathcal{L}_t^{\text{ELBO}} \cdot \mathbf{V}_t^\top)$
    
    \STATE Landing field updates:
    \STATE \quad $\boldsymbol{\Gamma}(\mathbf{U}_t) = \boldsymbol{\psi}(\mathbf{U}_t)\mathbf{U}_t + \lambda \cdot 4\mathbf{U}_t(\mathbf{U}_t^\top \mathbf{U}_t - \mathbf{I}_r)$
    \STATE \quad $\boldsymbol{\Gamma}(\mathbf{V}_t) = \boldsymbol{\psi}(\mathbf{V}_t)\mathbf{V}_t + \lambda \cdot 4\mathbf{V}_t(\mathbf{V}_t^\top \mathbf{V}_t - \mathbf{I}_r)$
    
    \STATE Update PoLAR parameters:
    \STATE \quad $\mathbf{U}_{t+1} = \mathbf{U}_t - \eta_{\text{polar}} \boldsymbol{\Gamma}(\mathbf{U}_t)$
    \STATE \quad $\mathbf{V}_{t+1} = \mathbf{V}_t - \eta_{\text{polar}} \boldsymbol{\Gamma}(\mathbf{V}_t)$
    \STATE \quad $\boldsymbol{\Lambda}_{t+1} = \boldsymbol{\Lambda}_t - \eta_{\text{polar}} \nabla_{\boldsymbol{\Lambda}} \mathcal{L}_t^{\text{ELBO}}$
    
    \STATE // VBLL parameter updates
    \FOR{$c = 1, \ldots, C$}
        \STATE Compute variational gradients using Eqs.~\ref{eq:grad_mu}--\ref{eq:grad_S}
        \STATE $\boldsymbol{\mu}_{c,t+1} = \boldsymbol{\mu}_{c,t} - \eta_{\text{vbll}} \frac{\partial \mathcal{L}_t^{\text{ELBO}}}{\partial \boldsymbol{\mu}_c}$
        \STATE $\mathbf{S}_{c,t+1} = \mathbf{S}_{c,t} - \eta_{\text{vbll}} \frac{\partial \mathcal{L}_t^{\text{ELBO}}}{\partial \mathbf{S}_c}$
        \STATE Project $\mathbf{S}_{c,t+1}$ to positive definite cone if necessary
    \ENDFOR
\ENDFOR
\STATE \textbf{Return} $\hat{\mathbf{U}} = \mathbf{U}_{T}$, $\hat{\boldsymbol{\Lambda}} = \boldsymbol{\Lambda}_{T}$, $\hat{\mathbf{V}} = \mathbf{V}_{T}$, $q(\boldsymbol{\Theta}) = \prod_{c=1}^C \mathcal{N}(\boldsymbol{\theta}_c; \boldsymbol{\mu}_{c,T}, \mathbf{S}_{c,T})$
\end{algorithmic}
\end{algorithm}

\begin{algorithm}[t]
\caption{Uncertainty-Aware Prediction in PoLAR-VBLL}
\label{alg:polar_vbll_inference}
\begin{algorithmic}[1]
\REQUIRE Converged PoLAR parameters $\{\hat{\mathbf{U}}, \hat{\boldsymbol{\Lambda}}, \hat{\mathbf{V}}\}$, variational posterior $q(\boldsymbol{\Theta})$, test input $\mathbf{x}^*$, training data $\mathcal{D}$, number of samples $K$, Laplace refinement flag
\ENSURE Predictive distribution $p(y^*|\mathbf{x}^*, \mathcal{D})$

\IF{Laplace refinement enabled}
    \STATE // Optional posterior refinement via Laplace approximation
    \STATE Compute converged means: $\hat{\boldsymbol{\mu}}_c = \boldsymbol{\mu}_{c,T}$ for $c = 1, \ldots, C$
    \STATE Compute Hessian using KFAC approximation:
    \STATE \quad $\mathbf{H}_c = -\nabla^2_{\boldsymbol{\theta}_c} \left( \log p(\mathcal{D} | \boldsymbol{\theta}_c, \hat{\boldsymbol{\Psi}}_{\text{polar}}) + \log p(\boldsymbol{\theta}_c) \right)\Big\rvert_{\boldsymbol{\theta}_c = \boldsymbol{\mu}_c^*}$
    \STATE Form Laplace posterior: $\tilde{q}(\boldsymbol{\theta}_c) = \mathcal{N}(\boldsymbol{\theta}_c; \hat{\boldsymbol{\mu}}_c, \boldsymbol{\Sigma}_c)$ where $\boldsymbol{\Sigma}_c = \mathbf{H}_c^{-1}$, for $c = 1, \ldots, C$
\ENDIF

\STATE // Extract test features
\STATE $\boldsymbol{\phi}^* = \boldsymbol{\phi}_{\mathbf{W}_0 + \hat{\mathbf{U}}\hat{\boldsymbol{\Lambda}}\hat{\mathbf{V}}^\top}(\mathbf{x}^*)$

\STATE // Monte Carlo sampling for predictive distribution
\STATE Initialize prediction accumulator: $\mathbf{p}_{\text{pred}} = \mathbf{0}_C$
\FOR{$k = 1, \ldots, K$}
    \IF{Laplace refinement enabled}
        \STATE Sample classification weights: $\boldsymbol{\theta}_c^{(k)} \sim q_{\text{Lap}}(\boldsymbol{\theta}_c)$ for $c = 1, \ldots, C$
    \ELSE
        \STATE Sample classification weights: $\boldsymbol{\theta}_c^{(k)} \sim q(\boldsymbol{\theta}_c) = \mathcal{N}(\boldsymbol{\mu}_{c,T}, \mathbf{S}_{c,T})$ for $c = 1, \ldots, C$
    \ENDIF
    \STATE Form weight matrix: $\boldsymbol{\Theta}^{(k)} = [\boldsymbol{\theta}_1^{(k)}, \ldots, \boldsymbol{\theta}_C^{(k)}]^\top$
    \STATE Compute logits: $\mathbf{z}^{(k)} = \boldsymbol{\Theta}^{(k)}\boldsymbol{\phi}^*$
    \STATE Compute sample prediction: $\mathbf{p}^{(k)} = \text{softmax}(\mathbf{z}^{(k)})$
    \STATE Accumulate: $\mathbf{p}_{\text{pred}} = \mathbf{p}_{\text{pred}} + \mathbf{p}^{(k)}$
\ENDFOR

\STATE Average predictions: $p(y^*|\mathbf{x}^*, \mathcal{D}) = \frac{1}{K}\mathbf{p}_{\text{pred}}$
\STATE \textbf{Return} Predictive distribution $p(y^*|\mathbf{x}^*, \mathcal{D})$
\end{algorithmic}
\end{algorithm}

\subsection{Detailed Gradient Derivations}
\label{sec:detailed_gradients}

This section provides the complete derivation of gradient updates for the joint PoLAR-VBLL optimization procedure described in Section~3.2.

\subsubsection{Variational Parameter Updates}

The gradients with respect to the variational parameters $\{\boldsymbol{\mu}_c, \mathbf{S}_c\}$ follow standard variational inference procedures:

\begin{align}
\frac{\partial \mathcal{L}^{\text{ELBO}}}{\partial \boldsymbol{\mu}_c} &= \sum_{n=1}^{N} \left[ y_{n,c} \boldsymbol{\phi}_{\mathbf{W}}(\mathbf{x}_n) - \frac{\exp(\boldsymbol{\mu}_c^\top \boldsymbol{\phi}_{\mathbf{W}}(\mathbf{x}_n) + \frac{1}{2}\boldsymbol{\phi}_{\mathbf{W}}(\mathbf{x}_n)^\top \mathbf{S}_c \boldsymbol{\phi}_{\mathbf{W}}(\mathbf{x}_n))}{\sum_{j=1}^C \exp(\boldsymbol{\mu}_j^\top \boldsymbol{\phi}_{\mathbf{W}}(\mathbf{x}_n) + \frac{1}{2}\boldsymbol{\phi}_{\mathbf{W}}(\mathbf{x}_n)^\top \mathbf{S}_j \boldsymbol{\phi}_{\mathbf{W}}(\mathbf{x}_n))} \boldsymbol{\phi}_{\mathbf{W}}(\mathbf{x}_n) \right] \nonumber\\
&\quad - \frac{1}{\sigma_\theta^2} \boldsymbol{\mu}_c \label{eq:grad_mu}
\end{align}

\begin{align}
\frac{\partial \mathcal{L}^{\text{ELBO}}}{\partial \mathbf{S}_c} &= \frac{1}{2} \sum_{n=1}^{N} \left[ - \frac{\exp(\boldsymbol{\mu}_c^\top \boldsymbol{\phi}_{\mathbf{W}}(\mathbf{x}_n) + \frac{1}{2}\boldsymbol{\phi}_{\mathbf{W}}(\mathbf{x}_n)^\top \mathbf{S}_c \boldsymbol{\phi}_{\mathbf{W}}(\mathbf{x}_n))}{\sum_{j=1}^C \exp(\boldsymbol{\mu}_j^\top \boldsymbol{\phi}_{\mathbf{W}}(\mathbf{x}_n) + \frac{1}{2}\boldsymbol{\phi}_{\mathbf{W}}(\mathbf{x}_n)^\top \mathbf{S}_j \boldsymbol{\phi}_{\mathbf{W}}(\mathbf{x}_n))} \boldsymbol{\phi}_{\mathbf{W}}(\mathbf{x}_n) \boldsymbol{\phi}_{\mathbf{W}}(\mathbf{x}_n)^\top \right] \nonumber\\
&\quad - \frac{1}{2\sigma_\theta^2} \mathbf{I}_d + \frac{1}{2} \mathbf{S}_c^{-1} \label{eq:grad_S}
\end{align}

\subsubsection{PoLAR Parameter Updates}

For the PoLAR parameters, we employ the chain rule to propagate gradients through the feature extractor $\boldsymbol{\phi}_{\mathbf{W}}(\mathbf{x})$ where $\mathbf{W} = \mathbf{W}_0 + \mathbf{U}\boldsymbol{\Lambda} \mathbf{V}^\top$. Let $\mathbf{G} := \frac{\partial \mathcal{L}^{\text{ELBO}}}{\partial (\mathbf{U}\boldsymbol{\Lambda} \mathbf{V}^\top)}$ denote the gradient with respect to the weight update. Then:

\begin{align}
\frac{\partial \mathcal{L}^{\text{ELBO}}}{\partial \boldsymbol{\Lambda}} &= \mathbf{U}^\top \mathbf{G} \mathbf{V} \label{eq:grad_Lambda}\\
\nabla_{\mathbf{U}} \mathcal{L}^{\text{ELBO}} &= \mathbf{G} \mathbf{V} \boldsymbol{\Lambda}^\top \label{eq:grad_U_euclidean}\\
\nabla_{\mathbf{V}} \mathcal{L}^{\text{ELBO}} &= \mathbf{G}^\top \mathbf{U} \boldsymbol{\Lambda} \label{eq:grad_V_euclidean}
\end{align}

\subsubsection{Riemannian Gradient Computation}

Since $\mathbf{U}$ and $\mathbf{V}$ are constrained to Stiefel manifolds, we convert the Euclidean gradients to their Riemannian counterparts. For a matrix $\mathbf{X} \in \text{St}(m, r)$, the Riemannian gradient is:

\begin{equation}
\text{grad}_R f(\mathbf{X}) = \nabla f(\mathbf{X}) - \mathbf{X} \nabla f(\mathbf{X})^\top \mathbf{X} \label{eq:riemannian_grad}
\end{equation}

Applying this to our PoLAR parameters:

\begin{align}
\boldsymbol{\psi}(\mathbf{U}) &= \text{Skew}(\nabla_{\mathbf{U}} \mathcal{L}^{\text{ELBO}} \cdot \mathbf{U}^\top) = \text{Skew}(\mathbf{G} \mathbf{V} \boldsymbol{\Lambda}^\top \mathbf{U}^\top) \label{eq:psi_U}\\
\boldsymbol{\psi}(\mathbf{V}) &= \text{Skew}(\nabla_{\mathbf{V}} \mathcal{L}^{\text{ELBO}} \cdot \mathbf{V}^\top) = \text{Skew}(\mathbf{G}^\top \mathbf{U} \boldsymbol{\Lambda} \mathbf{V}^\top) \label{eq:psi_V}
\end{align}

where $\text{Skew}(\mathbf{A}) = \frac{1}{2}(\mathbf{A} - \mathbf{A}^\top)$ extracts the skew-symmetric component.

\subsubsection{Landing Field Updates}

Following the infeasible optimization approach, we replace the expensive retraction operations with landing field updates:

\begin{align}
\boldsymbol{\Gamma}(\mathbf{U}) &= \boldsymbol{\psi}(\mathbf{U})\mathbf{U} + \lambda \nabla N(\mathbf{U}) \label{eq:landing_U}\\
\boldsymbol{\Gamma}(\mathbf{V}) &= \boldsymbol{\psi}(\mathbf{V})\mathbf{V} + \lambda \nabla N(\mathbf{V}) \label{eq:landing_V}
\end{align}

where $\nabla N(\mathbf{U}) = 4\mathbf{U}(\mathbf{U}^\top \mathbf{U} - \mathbf{I}_r)$ and $\nabla N(\mathbf{V}) = 4\mathbf{V}(\mathbf{V}^\top \mathbf{V} - \mathbf{I}_r)$ are the gradients of the infeasibility penalties $N(\mathbf{U}) = \|\mathbf{U}^\top \mathbf{U} - \mathbf{I}_r\|_F^2$ and $N(\mathbf{V}) = \|\mathbf{V}^\top \mathbf{V} - \mathbf{I}_r\|_F^2$, respectively.

The complete update procedure alternates between updating the variational parameters using standard gradient-based optimizers (e.g., Adam) on Eqs.~\ref{eq:grad_mu}--\ref{eq:grad_S}, and updating the PoLAR parameters using the landing field approach on Eqs.~\ref{eq:grad_Lambda}, \ref{eq:landing_U}, and \ref{eq:landing_V}.

\section{Implementation Details}
\label{sec:Implementation Details}

\subsection{Training Settings}

\paragraph{Model Architecture.} Our implementation builds upon the \texttt{LlaMA-3.1-8B} and \texttt{LLaMA-2-7B} foundation models~\citep{touvron2023llama}, utilizing its pre-trained language modeling head for VBLL mean initialization.

\paragraph{PoLAR Configuration.} The manifold penalty coefficient in PoLAR $\lambda = 1.0$. We parameterize the $\mathbf{S}$ matrix using the identity initialization and apply Landing Field optimization~\citep{zhang2025polar,gao2022optimization,schechtman2023orthogonal} with gradient type set to ''landing''. The Landing Field callback is enabled during training to maintain stability in optimization on the Grassmann manifold.

\paragraph{VBLL Parameterization.} For VBLL, we adopt the dense parameterization for computational efficiency while maintaining uncertainty quantification capabilities. The Jensen bound is used for approximating the softmax function. Prior hyperparameters are set as follows: prior scale $\sigma_\theta^2 = 1.0$, Wishart scale $\nu_0 = 10^{-2}$, degrees of freedom $\rho = 1.0$.
The regularization weight for KL divergence is computed as $\lambda_{\text{reg}} = 1/N$ where $N$ is the training set size. This regularization weight can be used to adjust the emphasis of model performance on ACC or Uncertainty Quantification ability. All parameter values are the default classification setting in the VBLL library~\citep{harrison2024variational}. 


\paragraph{Training Configuration.} 
For all shared parameters, we follow the settings of BLoB's official single-GPU scripts, except for LoRA Rank and LoRA Alpha, to ease BNN training and improve performance.
We train all methods (PoLAR-VBLL and baselines) for 5000 training steps with a batch size of 4, evaluation batch size of 8, and maximum sequence length of 300 tokens. All methods use AdamW~\citep{loshchilov2017decoupled} optimizers with learning rate $10^{-4}$ and a CosineAnnealingWarmRestarts scheduler~\citep{loshchilov2016sgdr}. 
Baselines are reproduced strictly according to the implementations in their official repositories. For sampling-based methods (BLoB, TFB, ScalaBL, C-LoRA), we set training sampling $K_{\text{train}}=1$ (single sample per forward pass) and inference sampling $K_{\text{eval}}=10$. LoRA/PoLAR rank ($r=8$), alpha ($\alpha=16$), and dropout (0.0). All training is conducted in BF16 precision on CUDA devices. For all MC-based uncertainty quantification evaluations, we use $n_{\text{samples}}=10$.


\paragraph{LA Calibration.} For post-hoc calibration, we apply LA with a diagonal Hessian structure over all model parameters. The prior precision is set to $1.0$.

\paragraph{Implementation of PoLAR-ized Variants.}
We adapt existing Bayesian LoRA methods to PoLAR by applying uncertainty quantification specifically to the core matrix $\boldsymbol{\Lambda}$ while preserving the Stiefel manifold constraints on the orthogonal factors $\mathbf{U}$ and $\mathbf{V}$.

\textbf{PoLAR-BLoB} replaces standard LoRA with the PoLAR parameterization $\Delta \mathbf{W} = \mathbf{U}\boldsymbol{\Lambda}\mathbf{V}^\top$ and applies variational inference to $\boldsymbol{\Lambda} \in \mathbb{R}^{r \times r}$, learning a diagonal covariance for the approximate posterior $q(\boldsymbol{\Lambda} | \mathcal{D})$ during training while keeping $\mathbf{U} \in \text{St}(m, r)$ and $\mathbf{V} \in \text{St}(n, r)$ deterministic.

\textbf{PoLAR-LA} performs MAP estimation on all PoLAR parameters $\{\mathbf{U}, \boldsymbol{\Lambda}, \mathbf{V}\}$, then computes the Laplace approximation by estimating the Hessian with respect to the adapter parameters at the MAP solution. \textbf{PoLAR-LA-LL} applies LA only to the last (classification) layer for direct comparison with our VBLL framework.

\textbf{PoLAR-TFB} applies training-free Bayesianization~\citep{shi2024training} by performing SVD on the trained core matrix $\boldsymbol{\Lambda}$ and inferring the posterior covariance scale $\beta$ via binary search on a held-out anchor set. However, despite our best efforts, we cannot find a proper set of hyperparameters to yield competitive performance.
We hypothesize that TFB's covariance estimation, originally designed for the full low-rank factors in standard LoRA, may not transfer effectively to the compact $r \times r$ core matrix in PoLAR. We therefore omit PoLAR-TFB and PoLAR-TFB-LL from our comparisons.

\subsection{Computational Environment}

\paragraph{Hardware Specifications.} All experiments are conducted on a high-performance computing system equipped with NVIDIA RTX A6000 Ada GPUs and AMD 9600 Threadripper processors with 64 cores and 128 threads. This configuration provides substantial computational resources for both GPU-accelerated training and CPU-intensive operations such as Hessian computation for Laplace approximation.

\paragraph{Software Dependencies.} Our implementation leverages several key Python packages: PyTorch~\citep{paszke2019pytorch} for deep learning operations, HuggingFace PEFT~\citep{mangrulkar2022peft} for adapter implementations, custom Laplace approximation libraries~\citep{yang2023bayesian, laplace2021, kristiadi2024sober} for post-hoc uncertainty calibration, PoLAR optimization libraries~\citep{zhang2025polar}, and VBLL (Variational Bayesian Last Layer) implementations~\citep{harrison2024variational}. Complete dependency specifications and version information are provided in our \texttt{requirements.txt} file, which will be made available upon acceptance.

\section{Extend Experiments}
\label{sec: Additional Experimental Results}


\subsection{Additional Backbone Evaluation on LlaMA-2-7B}
\label{sec: Full comparison}

We conducted comprehensive experiments on \texttt{LlaMA-2-7B} across four datasets with all baseline methods. As shown in Table~\ref{tab:main_results_full}, BLoB exhibits a trade-off between accuracy and uncertainty quantification: at $N=0$ sampling, it achieves higher accuracy but with reduced calibration quality; as sampling increases to $N=10$, BLoB shows improved uncertainty metrics but with degraded predictive accuracy. In contrast, our PoLAR-VBLL framework achieves competitive accuracy across all datasets while maintaining strong uncertainty calibration, demonstrating that high predictive performance and well-calibrated uncertainties can be obtained simultaneously without requiring the typical trade-off. The optional LA refinement further enhances ECE and NLL while preserving accuracy, confirming that our variational training provides a robust foundation for posterior refinement. These results demonstrate that our framework generalizes effectively beyond \texttt{LlaMA-3.1-8B}, consistently achieving superior uncertainty quantification while maintaining competitive predictive performance.

\begin{table}[t]
\centering
\caption{Performances on ID datasets in terms of ACC, ECE, and NLL using \texttt{LlaMA2-7B}. Bold and underlined denote the best and the second-best performance, respectively. Here, we include PoLAR-VBLL with and without LA.}
\label{tab:main_results_full}
\resizebox{\textwidth}{!}{%
\begin{tabular}{ll|cccc}
\toprule
\multirow{2}{*}{\textbf{Metric}} & \multirow{2}{*}{\textbf{Method}} & \multicolumn{4}{c}{\textbf{Datasets}} \\
\cmidrule{3-6}
& & \textbf{WG-S} & \textbf{ARC-C} & \textbf{ARC-E} & \textbf{OBQA} \\
\midrule
\multirow{11}{*}{\textbf{ACC (\%)}} 
& MLE & 68.99$\pm$0.58 & 69.10$\pm$2.84 & 85.65$\pm$0.92 & 81.52$\pm$0.25 \\
& MAP & 68.62$\pm$0.71 & 67.59$\pm$0.40 & 86.55$\pm$0.55 & 81.38$\pm$0.65 \\
& MCD & 69.46$\pm$0.62 & 68.69$\pm$1.30 & 86.21$\pm$0.46 & 81.72$\pm$0.10 \\
& ENS & 69.57$\pm$0.66 & 66.20$\pm$2.01 & 84.40$\pm$0.81 & 81.38$\pm$0.91 \\
& BBB & 66.54$\pm$7.87 & 68.13$\pm$1.27 & \underline{86.86$\pm$0.74} & 82.06$\pm$0.59 \\
& LA & 69.45$\pm$1.73 & 66.78$\pm$0.69 & 80.05$\pm$0.22 & 82.07$\pm$0.67 \\
& BLoB ($N$=0) & \underline{70.89$\pm$0.82} & \underline{70.83$\pm$1.57} & 86.68$\pm$0.60 & \textbf{82.73$\pm$0.41} \\
& BLoB ($N$=10) & 69.07$\pm$0.34 & 68.81$\pm$1.09 & 86.56$\pm$0.35 & 81.52$\pm$0.74 \\
\cmidrule{2-6}
& PoLAR-VBLL (wo. LA)& \textbf{71.62$\pm$0.27} & \textbf{70.92$\pm$0.24} & \textbf{88.03$\pm$0.44} & \underline{82.53$\pm$0.12} \\
& \textbf{PoLAR-VBLL} & \textbf{71.62$\pm$0.27} & \textbf{70.92$\pm$0.24} & \textbf{88.03$\pm$0.44} & \underline{82.53$\pm$0.12} \\
\midrule
\multirow{11}{*}{\textbf{ECE (\%)}}
& MLE & 29.83$\pm$0.58 & 29.00$\pm$1.97 & 13.12$\pm$1.39 & 12.55$\pm$0.46 \\
& MAP & 29.76$\pm$0.87 & 29.42$\pm$0.68 & 12.07$\pm$0.55 & 13.26$\pm$0.82 \\
& MCD & 27.98$\pm$0.44 & 27.53$\pm$0.80 & 12.20$\pm$0.56 & 13.10$\pm$0.11 \\
& ENS & 28.52$\pm$0.55 & 29.16$\pm$2.37 & 12.57$\pm$0.58 & 15.34$\pm$0.27 \\
& BBB & 21.81$\pm$12.95 & 26.23$\pm$1.47 & 12.28$\pm$0.58 & 11.38$\pm$1.07 \\
& LA & 13.47$\pm$1.43 & 16.25$\pm$2.61 & 33.29$\pm$0.57 & 6.12$\pm$1.55 \\
& BLoB ($N$=0) & 20.62$\pm$0.83 & 20.61$\pm$1.16 & 9.43$\pm$0.38 & 8.36$\pm$0.38 \\
& BLoB ($N$=10) & 9.35$\pm$1.37 & 9.59$\pm$1.88 & \underline{3.64$\pm$0.53} & \textbf{3.77$\pm$1.47} \\
\cmidrule{2-6}
& PoLAR-VBLL (wo. LA) & \underline{8.26$\pm$0.60} & \underline{8.36$\pm$0.13} & 5.22$\pm$0.41 & 5.58$\pm$0.34 \\
& \textbf{PoLAR-VBLL} & \textbf{7.31$\pm$0.32} & \textbf{7.41$\pm$0.78} & \textbf{2.63$\pm$0.81} & \underline{4.63$\pm$1.43} \\
\midrule
\multirow{11}{*}{\textbf{NLL}}
& MLE & 3.17$\pm$0.37 & 2.85$\pm$0.27 & 1.17$\pm$0.13 & 0.73$\pm$0.03 \\
& MAP & 2.46$\pm$0.34 & 2.66$\pm$0.11 & 0.90$\pm$0.05 & 0.75$\pm$0.01 \\
& MCD & 2.79$\pm$0.53 & 2.67$\pm$0.15 & 1.00$\pm$0.14 & 0.77$\pm$0.03 \\
& ENS & 2.71$\pm$0.08 & 2.46$\pm$0.22 & 0.82$\pm$0.03 & 1.06$\pm$0.04 \\
& BBB & 1.40$\pm$0.55 & 2.23$\pm$0.04 & 0.91$\pm$0.06 & 0.66$\pm$0.05 \\
& LA & 0.67$\pm$0.01 & 1.03$\pm$0.04 & 0.88$\pm$0.00 & 0.72$\pm$0.01 \\
& BLoB ($N$=0) & 0.91$\pm$0.10 & 1.19$\pm$0.02 & 0.56$\pm$0.01 & \underline{0.56$\pm$0.02} \\
& BLoB ($N$=10) & \underline{0.63$\pm$0.01} & \textbf{0.78$\pm$0.02} & \textbf{0.40$\pm$0.01} & \textbf{0.50$\pm$0.01} \\
\cmidrule{2-6}
& PoLAR-VBLL (wo. LA) & 0.66$\pm$0.03 & 0.95$\pm$0.07 & 0.51$\pm$0.03 & 0.64$\pm$0.01\\
& \textbf{PoLAR-VBLL} & \textbf{0.60$\pm$0.01} & \underline{0.91$\pm$0.00} & \underline{0.47$\pm$0.03} & 0.63$\pm$0.02 \\
\bottomrule
\end{tabular}%
}
\end{table}

\subsection{Memory Usage and Runtime}
\label{sec: Memory usage and Run time}

We evaluate the computational efficiency of different uncertainty quantification methods on the ARC-E dataset. All experiments use a training batch size of 4, inference batch size of 4, LoRA rank $r=8$, $\alpha=16$, and sequence length of 300.

\begin{table}[h]
\centering
\caption{GPU memory usage and inference time comparison across different uncertainty quantification methods. \\ \textbf{Configuration:} LoRA rank = 8, batch size = 4, dataset = ARC-E, LoRA targets = [\texttt{q\_proj}, \texttt{v\_proj}].}
\label{tab:uncertainty_ablation_GPU}
\begin{tabular}{l|ccc}
\toprule
\textbf{Method} & \textbf{Training Memory (MB)} & \textbf{Test Memory (MB)} & \textbf{Inference Time (s)} \\
\midrule
MLE & 19,474 & 17,298 & 8 \\
PoLAR-MLE & 19,556 & 17,150 & 7 \\
PoLAR-VBLL (Ours) & 20,178 & 18,423 & 12 \\
PoLAR-BLoB & 19,738 & 19,830 & 80 \\
LoRA-BLoB & 20,972 & 18,762 & 89 \\
PoLAR-LA-LL & 19,556 & 18,074 & 30 \\
PoLAR-LA & 19,556 & 40,737 & 38 \\
LoRA-LA-LL & 19,474 & 16,313 & 32 \\
LoRA-LA & 19,474 & 42,766 & 40 \\
LoRA-VBLL & 19,560 & 17,764 & 8 \\
TFB & 20,972 & 24,432 & 90 \\
TFB-LL & 20,972 & 22,306 & 80 \\
ScalaBL & 17,290 & 16,552 & 52 \\
C-LoRA & 18,338 & 17,102 & 60 \\
\bottomrule
\end{tabular}
\end{table}

As shown in Table~\ref{tab:uncertainty_ablation_GPU}, PoLAR-VBLL achieves approximately $7\times$ inference speedup compared to BLoB-based methods (12s vs.\ 80--90s) while maintaining a competitive memory footprint significantly lower than full-network Laplace approximations (18,423 MB vs.\ $\sim$41,000 MB for PoLAR-LA and LoRA-LA).

The efficiency of PoLAR-VBLL stems from two key design choices. First, while BLoB-based methods and TFB require multiple complete forward passes through the entire LLM backbone for uncertainty estimation, our framework employs head-only sampling only requires a single backbone pass followed by lightweight sampling at the last layer. Second, our analytical Jenson-bounded ELBO formulation eliminates the sampling overhead inherent in Monte Carlo-based training, whereas BLoB incurs approximately 50\% additional parameter overhead~\citep{samplawski2025scalable, rahmati2025c} by maintaining both mean and variance parameters across all adapter layers.

Table~\ref{tab:variational_params} details the variational parameter counts per layer, illustrating the source of memory differences among variational methods.

\begin{table}[h]
\centering
\caption{Variational parameters per layer for different methods.}
\label{tab:variational_params}
\begin{tabular}{l|cc}
\toprule
\textbf{Method} & \textbf{Variational Parameters per Layer} & \textbf{Calculation} \\
\midrule
LoRA-BLoB & 131,072 & $r \times d \times 2 = 16 \times 4096 \times 2$ \\
PoLAR-BLoB & 512 & $r \times r \times 2 = 16 \times 16 \times 2$ \\
ScalaBL & 32 & $r \times 2 = 16 \times 2$ \\
\bottomrule
\end{tabular}
\end{table}

The $256\times$ reduction in variational parameters from LoRA-BLoB to PoLAR-BLoB arises because LoRA-BLoB performs variational inference on full low-rank matrices of dimension $d \times r$, whereas PoLAR-BLoB restricts stochasticity to the compact core matrix of dimension $r \times r$. C-LoRA achieves lower memory by using deterministic contextual MLPs instead of variational parameters, avoiding both reparameterization overhead and doubled optimizer states.

In summary, PoLAR-VBLL bridges predictive performance and computational efficiency, making it practical for resource-constrained deployment scenarios.

\begin{table*}[t]
\centering
\caption{Performance of different methods on \texttt{Llama-3.1-8B}. ACC and ECE are reported in percentages. The evaluation is done across six in-distribution common-sense reasoning datasets with fine-tuning of 5000 steps. For OOD evaluation, models are trained on OBQA and tested on other datasets. \textbf{Bold} and \underline{underlined} denote the best and second-best performance, respectively.}
\label{tab:llama3_results_full}

\setlength{\tabcolsep}{2.5pt}
\resizebox{\textwidth}{!}{%
\scriptsize
\begin{tabular}{ll|cccccc|cc|cc}
\toprule
& & \multicolumn{6}{c|}{\textbf{In-Distribution Datasets}} & \multicolumn{4}{c}{\textbf{Out-of-Distribution Datasets} (OBQA$\rightarrow$X)} \\
\cmidrule{3-8} \cmidrule{9-12}
& & & & & & & & \multicolumn{2}{c|}{\textit{Small Shift}} & \multicolumn{2}{c}{\textit{Large Shift}} \\
\textbf{Metric} & \textbf{Method} & \textbf{WG-S} & \textbf{ARC-C} & \textbf{ARC-E} & \textbf{WG-M} & \textbf{OBQA} & \textbf{BoolQ} & \textbf{ARC-C} & \textbf{ARC-E} & \textbf{Chem} & \textbf{Phy} \\
\midrule
\multirow{17}{*}{\rotatebox{90}{\textbf{ACC ($\uparrow$)}}}
& MLE & \underline{77.92$_{\pm0.62}$} & 81.05$_{\pm1.62}$ & 90.66$_{\pm0.10}$ & 82.80$_{\pm0.96}$ & \underline{88.30$_{\pm0.36}$} & 87.86$_{\pm0.50}$ & 79.33$_{\pm0.64}$ & 85.66$_{\pm0.50}$ & 48.00$_{\pm2.00}$ & 43.33$_{\pm1.53}$ \\
& LA & \underline{77.92$_{\pm0.62}$} & 81.05$_{\pm1.62}$ & 90.66$_{\pm0.10}$ & 82.80$_{\pm0.96}$ & \underline{88.30$_{\pm0.36}$} & 87.86$_{\pm0.50}$ & 79.33$_{\pm0.64}$ & 85.66$_{\pm0.50}$ & 48.00$_{\pm2.00}$ & 43.33$_{\pm1.53}$ \\
& MAP & 77.32$_{\pm1.31}$ & 80.59$_{\pm0.86}$ & 90.01$_{\pm0.37}$ & 81.96$_{\pm0.36}$ & 88.13$_{\pm0.64}$ & 88.54$_{\pm0.34}$ & 79.50$_{\pm0.56}$ & 85.29$_{\pm0.61}$ & 47.33$_{\pm1.15}$ & 44.00$_{\pm5.00}$ \\
& MCD & 76.92$_{\pm0.92}$ & 81.14$_{\pm1.50}$ & 91.01$_{\pm0.40}$ & 82.90$_{\pm0.39}$ & 88.00$_{\pm0.20}$ & 88.62$_{\pm0.15}$ & 80.33$_{\pm0.52}$ & 84.78$_{\pm0.61}$ & 46.67$_{\pm2.89}$ & 43.67$_{\pm2.08}$ \\
& ENS & 77.00$_{\pm0.25}$ & \underline{81.74$_{\pm1.33}$} & 90.81$_{\pm1.00}$ & 82.67$_{\pm0.67}$ & 88.20$_{\pm0.20}$ & 88.05$_{\pm0.92}$ & 79.16$_{\pm0.70}$ & 83.99$_{\pm0.30}$ & 45.00$_{\pm1.00}$ & 44.33$_{\pm2.52}$ \\
& PoLAR-MLE & \textbf{78.09$_{\pm0.39}$} & 81.21$_{\pm1.02}$ & 90.25$_{\pm1.24}$ & \textbf{83.61$_{\pm0.32}$} & 86.67$_{\pm1.22}$ & \textbf{89.16$_{\pm0.27}$} & \underline{80.97$_{\pm0.76}$} & 85.57$_{\pm0.21}$ & \underline{48.33$_{\pm0.58}$} & 43.33$_{\pm2.52}$ \\
& PoLAR-LA & \textbf{78.09$_{\pm0.39}$} & 81.21$_{\pm1.02}$ & 90.25$_{\pm1.24}$ & \textbf{83.61$_{\pm0.32}$} & 86.67$_{\pm1.22}$ & \textbf{89.16$_{\pm0.27}$} & \underline{80.97$_{\pm0.76}$} & 85.57$_{\pm0.21}$ & \underline{48.33$_{\pm0.58}$} & 43.33$_{\pm2.52}$ \\
& PoLAR-LA-LL & \textbf{78.09$_{\pm0.39}$} & 81.21$_{\pm1.02}$ & 90.25$_{\pm1.24}$ & \textbf{83.61$_{\pm0.32}$} & 86.67$_{\pm1.22}$ & \textbf{89.16$_{\pm0.27}$} & \underline{80.97$_{\pm0.76}$} & 85.57$_{\pm0.21}$ & \underline{48.33$_{\pm0.58}$} & 43.33$_{\pm2.52}$ \\
& TFB-LL & 76.25$_{\pm0.63}$ & 80.58$_{\pm1.10}$ & 91.03$_{\pm0.72}$ & 82.70$_{\pm0.33}$ & \underline{88.30$_{\pm0.56}$} & 87.53$_{\pm0.62}$ & \underline{80.97$_{\pm1.70}$} & \underline{85.74$_{\pm0.47}$} & 47.33$_{\pm2.08}$ & 45.67$_{\pm0.58}$ \\
& TFB & 73.53$_{\pm0.87}$ & 80.31$_{\pm1.15}$ & \underline{91.20$_{\pm1.40}$} & 80.71$_{\pm0.44}$ & 86.80$_{\pm1.06}$ & 87.83$_{\pm0.72}$ & 80.90$_{\pm1.51}$ & 84.50$_{\pm1.03}$ & 46.87$_{\pm1.04}$ & \underline{48.83$_{\pm1.92}$} \\
& C-LoRA & 77.16$_{\pm0.58}$ & 78.95$_{\pm0.53}$ & 90.40$_{\pm1.10}$ & 82.16$_{\pm0.32}$ & 86.83$_{\pm0.43}$ & 88.26$_{\pm0.92}$ & 80.07$_{\pm1.79}$ & 85.04$_{\pm0.36}$ & 47.67$_{\pm3.06}$ & 41.33$_{\pm2.89}$ \\
& BLoB & 72.36$_{\pm0.96}$ & 79.42$_{\pm1.19}$ & 90.16$_{\pm1.07}$ & 79.32$_{\pm0.95}$ & 87.53$_{\pm1.17}$ & 87.54$_{\pm0.54}$ & 79.10$_{\pm0.91}$ & 84.20$_{\pm1.01}$ & 45.67$_{\pm4.51}$ & 45.67$_{\pm0.58}$ \\
& PoLAR-BLoB & 76.49$_{\pm0.34}$ & 80.03$_{\pm1.59}$ & 91.19$_{\pm0.27}$ & 82.29$_{\pm0.37}$ & 87.67$_{\pm0.46}$ & 87.73$_{\pm0.82}$ & 80.34$_{\pm1.33}$ & 84.29$_{\pm1.08}$ & 46.18$_{\pm3.18}$ & 46.88$_{\pm2.09}$ \\
\cmidrule{2-12}
& PoLAR-VBLL (w/o LA) & 77.26$_{\pm0.50}$ & \textbf{81.79$_{\pm0.42}$} & \textbf{91.38$_{\pm0.39}$} & \underline{83.04$_{\pm0.46}$} & \textbf{88.43$_{\pm0.25}$} & \underline{88.88$_{\pm0.43}$} & \textbf{81.11$_{\pm0.82}$} & \textbf{85.92$_{\pm0.50}$} & \textbf{49.30$_{\pm1.61}$} & \textbf{48.91$_{\pm1.05}$} \\
& \textbf{PoLAR-VBLL} & 77.26$_{\pm0.50}$ & \textbf{81.79$_{\pm0.42}$} & \textbf{91.38$_{\pm0.39}$} & \underline{83.04$_{\pm0.46}$} & \textbf{88.43$_{\pm0.25}$} & \underline{88.88$_{\pm0.43}$} & \textbf{81.11$_{\pm0.82}$} & \textbf{85.92$_{\pm0.50}$} & \textbf{49.30$_{\pm1.61}$} & \textbf{48.91$_{\pm1.05}$} \\
\midrule
\multirow{17}{*}{\rotatebox{90}{\textbf{ECE ($\downarrow$)}}}
& MLE & 21.11$_{\pm0.56}$ & 17.95$_{\pm1.83}$ & 8.95$_{\pm0.21}$ & 15.46$_{\pm0.71}$ & 8.33$_{\pm0.19}$ & 4.76$_{\pm0.60}$ & 13.89$_{\pm1.04}$ & 10.31$_{\pm1.06}$ & 28.73$_{\pm1.02}$ & 37.02$_{\pm2.77}$ \\
& LA & 16.41$_{\pm1.20}$ & 9.72$_{\pm1.28}$ & 4.51$_{\pm0.31}$ & 8.37$_{\pm0.82}$ & 7.40$_{\pm0.13}$ & 2.33$_{\pm0.42}$ & 5.85$_{\pm2.05}$ & 5.09$_{\pm0.17}$ & 11.69$_{\pm0.91}$ & 13.02$_{\pm2.87}$ \\
& MAP & 20.87$_{\pm1.75}$ & 18.03$_{\pm0.18}$ & 9.30$_{\pm0.27}$ & 15.82$_{\pm0.34}$ & 9.09$_{\pm0.42}$ & 4.51$_{\pm0.16}$ & 13.92$_{\pm1.83}$ & 10.22$_{\pm0.57}$ & 30.78$_{\pm2.18}$ & 36.55$_{\pm3.59}$ \\
& MCD & 21.58$_{\pm1.09}$ & 17.21$_{\pm1.54}$ & 8.13$_{\pm0.37}$ & 14.46$_{\pm0.35}$ & 9.28$_{\pm0.34}$ & 4.63$_{\pm0.11}$ & 12.79$_{\pm1.20}$ & 9.95$_{\pm0.75}$ & 30.04$_{\pm1.39}$ & 35.99$_{\pm2.68}$ \\
& ENS & 18.89$_{\pm1.97}$ & 15.62$_{\pm1.20}$ & 8.28$_{\pm0.39}$ & 13.87$_{\pm0.91}$ & 7.88$_{\pm0.54}$ & 3.61$_{\pm0.25}$ & 12.51$_{\pm1.19}$ & 12.64$_{\pm1.62}$ & 17.09$_{\pm2.97}$ & 23.10$_{\pm1.57}$ \\
& PoLAR-MLE & 20.48$_{\pm0.99}$ & 16.99$_{\pm1.77}$ & 9.02$_{\pm0.93}$ & 18.64$_{\pm1.21}$ & 8.56$_{\pm1.50}$ & 2.12$_{\pm0.21}$ & 10.84$_{\pm0.39}$ & 8.35$_{\pm0.63}$ & 26.33$_{\pm2.01}$ & 33.22$_{\pm1.27}$ \\
& PoLAR-LA & 15.19$_{\pm5.43}$ & 9.69$_{\pm1.06}$ & 6.15$_{\pm1.06}$ & \textbf{3.16$_{\pm0.65}$} & 6.04$_{\pm0.41}$ & \underline{1.88$_{\pm0.29}$} & 7.09$_{\pm1.63}$ & 5.19$_{\pm0.52}$ & \underline{11.48$_{\pm1.07}$} & 16.09$_{\pm3.10}$ \\
& PoLAR-LA-LL & 15.06$_{\pm9.29}$ & 8.36$_{\pm1.76}$ & 5.41$_{\pm0.10}$ & 8.30$_{\pm0.79}$ & 6.21$_{\pm0.69}$ & 1.96$_{\pm0.24}$ & 9.45$_{\pm0.83}$ & 7.27$_{\pm0.97}$ & 14.89$_{\pm1.12}$ & 13.75$_{\pm1.84}$ \\
& TFB-LL & 12.34$_{\pm0.36}$ & 10.43$_{\pm1.69}$ & 3.77$_{\pm0.12}$ & 8.02$_{\pm0.86}$ & 4.36$_{\pm0.53}$ & 3.13$_{\pm0.71}$ & 7.86$_{\pm1.69}$ & 5.35$_{\pm0.81}$ & 18.75$_{\pm4.23}$ & 20.67$_{\pm3.02}$ \\
& TFB & 5.90$_{\pm0.56}$ & \underline{4.96$_{\pm1.45}$} & \underline{3.72$_{\pm0.22}$} & \underline{3.28$_{\pm0.64}$} & 6.18$_{\pm0.43}$ & 4.21$_{\pm0.42}$ & 5.10$_{\pm0.37}$ & 4.03$_{\pm1.25}$ & 17.83$_{\pm2.71}$ & 15.80$_{\pm2.32}$ \\
& C-LoRA & 18.31$_{\pm0.42}$ & 8.13$_{\pm0.79}$ & 5.42$_{\pm0.16}$ & 6.22$_{\pm1.07}$ & 5.33$_{\pm0.75}$ & 3.63$_{\pm0.63}$ & 13.94$_{\pm1.99}$ & 10.38$_{\pm0.87}$ & 27.99$_{\pm4.78}$ & 35.25$_{\pm3.44}$ \\
& BLoB & 5.78$_{\pm0.75}$ & 7.34$_{\pm1.31}$ & 5.66$_{\pm0.65}$ & 3.91$_{\pm0.93}$ & 5.30$_{\pm1.72}$ & 2.61$_{\pm0.49}$ & \underline{4.87$_{\pm0.56}$} & 5.05$_{\pm0.57}$ & 13.23$_{\pm3.50}$ & \underline{11.31$_{\pm1.86}$} \\
& PoLAR-BLoB & \underline{5.21$_{\pm0.92}$} & 7.02$_{\pm0.96}$ & 5.76$_{\pm0.33}$ & 7.70$_{\pm1.07}$ & \underline{2.36$_{\pm0.69}$} & 2.38$_{\pm0.36}$ & 6.52$_{\pm0.67}$ & \textbf{3.76$_{\pm0.97}$} & 20.29$_{\pm3.49}$ & 18.43$_{\pm2.32}$ \\
\cmidrule{2-12}
& PoLAR-VBLL (w/o LA) & 8.19$_{\pm1.88}$ & 5.86$_{\pm0.49}$ & 4.90$_{\pm1.43}$ & 8.79$_{\pm0.66}$ & 2.96$_{\pm0.13}$ & 2.14$_{\pm0.19}$ & 7.65$_{\pm1.16}$ & 4.09$_{\pm0.94}$ & 15.05$_{\pm1.67}$ & 15.63$_{\pm1.90}$ \\
& \textbf{PoLAR-VBLL} & \textbf{3.89$_{\pm1.27}$} & \textbf{4.92$_{\pm0.49}$} & \textbf{3.71$_{\pm0.29}$} & 3.77$_{\pm0.53}$ & \textbf{2.34$_{\pm0.66}$} & \textbf{1.77$_{\pm0.50}$} & \textbf{4.55$_{\pm0.22}$} & \underline{3.89$_{\pm0.68}$} & \textbf{10.30$_{\pm1.36}$} & \textbf{11.12$_{\pm1.40}$} \\
\midrule
\multirow{17}{*}{\rotatebox{90}{\textbf{NLL ($\downarrow$)}}}
& MLE & 2.23$_{\pm0.01}$ & 1.75$_{\pm0.09}$ & 0.74$_{\pm0.06}$ & 1.05$_{\pm0.12}$ & 0.49$_{\pm0.01}$ & \underline{0.27$_{\pm0.01}$} & 0.90$_{\pm0.04}$ & 0.63$_{\pm0.06}$ & 1.75$_{\pm0.06}$ & 1.94$_{\pm0.08}$ \\
& LA & \textbf{0.57$_{\pm0.01}$} & 1.04$_{\pm0.02}$ & 0.61$_{\pm0.07}$ & 0.56$_{\pm0.06}$ & 0.39$_{\pm0.01}$ & \underline{0.27$_{\pm0.01}$} & \underline{0.54$_{\pm0.01}$} & \textbf{0.39$_{\pm0.02}$} & 1.19$_{\pm0.02}$ & 1.22$_{\pm0.03}$ \\
& MAP & 2.12$_{\pm0.23}$ & 1.91$_{\pm0.11}$ & 0.86$_{\pm0.19}$ & 0.89$_{\pm0.05}$ & 0.56$_{\pm0.03}$ & 0.28$_{\pm0.01}$ & 0.95$_{\pm0.10}$ & 0.64$_{\pm0.02}$ & 1.81$_{\pm0.09}$ & 1.95$_{\pm0.14}$ \\
& MCD & 2.24$_{\pm0.45}$ & 1.78$_{\pm0.08}$ & 0.82$_{\pm0.10}$ & 0.94$_{\pm0.07}$ & 0.55$_{\pm0.07}$ & 0.28$_{\pm0.00}$ & 0.95$_{\pm0.07}$ & 0.65$_{\pm0.07}$ & 1.84$_{\pm0.05}$ & 2.01$_{\pm0.13}$ \\
& ENS & 1.42$_{\pm0.13}$ & 1.34$_{\pm0.17}$ & 0.61$_{\pm0.10}$ & 0.68$_{\pm0.03}$ & 0.41$_{\pm0.01}$ & \underline{0.27$_{\pm0.01}$} & 0.99$_{\pm0.06}$ & 0.74$_{\pm0.04}$ & 1.41$_{\pm0.05}$ & 1.46$_{\pm0.06}$ \\
& PoLAR-MLE & 1.97$_{\pm0.06}$ & 1.02$_{\pm0.02}$ & 0.79$_{\pm0.03}$ & 0.90$_{\pm0.04}$ & 0.48$_{\pm0.05}$ & \underline{0.27$_{\pm0.00}$} & 0.68$_{\pm0.04}$ & 0.50$_{\pm0.04}$ & 1.56$_{\pm0.06}$ & 1.61$_{\pm0.04}$ \\
& PoLAR-LA & 0.61$_{\pm0.03}$ & 0.73$_{\pm0.07}$ & 0.39$_{\pm0.04}$ & \textbf{0.53$_{\pm0.01}$} & \underline{0.37$_{\pm0.07}$} & \underline{0.27$_{\pm0.01}$} & 0.56$_{\pm0.02}$ & 0.42$_{\pm0.03}$ & \underline{1.18$_{\pm0.04}$} & \textbf{1.20$_{\pm0.04}$} \\
& PoLAR-LA-LL & 0.76$_{\pm0.28}$ & 0.61$_{\pm0.04}$ & 0.36$_{\pm0.02}$ & \underline{0.55$_{\pm0.04}$} & 0.39$_{\pm0.03}$ & \underline{0.27$_{\pm0.00}$} & 0.57$_{\pm0.02}$ & 0.48$_{\pm0.03}$ & 1.42$_{\pm0.03}$ & 1.46$_{\pm0.05}$ \\
& TFB-LL & 0.59$_{\pm0.01}$ & 0.64$_{\pm0.04}$ & 0.35$_{\pm0.06}$ & 0.61$_{\pm0.01}$ & \textbf{0.35$_{\pm0.06}$} & \underline{0.27$_{\pm0.01}$} & 0.58$_{\pm0.03}$ & \textbf{0.39$_{\pm0.01}$} & 1.27$_{\pm0.06}$ & 1.37$_{\pm0.09}$ \\
& TFB & 0.59$_{\pm0.01}$ & \textbf{0.58$_{\pm0.07}$} & \underline{0.33$_{\pm0.03}$} & \underline{0.55$_{\pm0.03}$} & 0.39$_{\pm0.01}$ & \underline{0.27$_{\pm0.01}$} & \underline{0.54$_{\pm0.01}$} & 0.44$_{\pm0.11}$ & 1.23$_{\pm0.03}$ & 1.31$_{\pm0.04}$ \\
& C-LoRA & 0.85$_{\pm0.02}$ & 0.90$_{\pm0.07}$ & 0.35$_{\pm0.01}$ & 0.62$_{\pm0.05}$ & 0.58$_{\pm0.06}$ & 0.29$_{\pm0.02}$ & 0.83$_{\pm0.08}$ & 0.63$_{\pm0.04}$ & 1.69$_{\pm0.07}$ & 1.84$_{\pm0.10}$ \\
& BLoB & \underline{0.58$_{\pm0.01}$} & \underline{0.59$_{\pm0.03}$} & \textbf{0.30$_{\pm0.08}$} & 0.60$_{\pm0.05}$ & \underline{0.37$_{\pm0.01}$} & 0.31$_{\pm0.03}$ & 0.55$_{\pm0.02}$ & \underline{0.41$_{\pm0.01}$} & \underline{1.18$_{\pm0.03}$} & 1.31$_{\pm0.06}$ \\
& PoLAR-BLoB & 0.67$_{\pm0.06}$ & \underline{0.59$_{\pm0.04}$} & 0.35$_{\pm0.01}$ & 0.61$_{\pm0.04}$ & \textbf{0.35$_{\pm0.01}$} & 0.29$_{\pm0.01}$ & 0.55$_{\pm0.02}$ & \underline{0.41$_{\pm0.04}$} & 1.26$_{\pm0.05}$ & \underline{1.21$_{\pm0.03}$} \\
\cmidrule{2-12}
& PoLAR-VBLL (w/o LA) & 0.61$_{\pm0.02}$ & 0.64$_{\pm0.02}$ & 0.36$_{\pm0.01}$ & \underline{0.55$_{\pm0.02}$} & 0.41$_{\pm0.02}$ & 0.32$_{\pm0.03}$ & 0.56$_{\pm0.03}$ & 0.44$_{\pm0.01}$ & 1.21$_{\pm0.08}$ & 1.27$_{\pm0.05}$ \\
& \textbf{PoLAR-VBLL} & \underline{0.58$_{\pm0.01}$} & \textbf{0.58$_{\pm0.05}$} & \underline{0.33$_{\pm0.02}$} & \textbf{0.53$_{\pm0.02}$} & \textbf{0.35$_{\pm0.01}$} & \textbf{0.26$_{\pm0.02}$} & \textbf{0.53$_{\pm0.02}$} & 0.42$_{\pm0.02}$ & \textbf{1.16$_{\pm0.01}$} & \textbf{1.20$_{\pm0.05}$} \\
\bottomrule
\end{tabular}%
}
\end{table*}

\subsection{Ablation Study: Disentangling PoLAR, LA and VBLL Contributions}
\label{sec:ablation_blob}

To isolate the contribution of each component in our framework, we systematically compare different uncertainty quantification methods under identical PoLAR adapters. Since our method applies Laplace Approximation (LA) exclusively to the last layer, we include PoLAR-LA-LL (which applies LA only to the last layer of a deterministically trained model) alongside PoLAR-LA (which applies LA across all adapter layers) to ensure fair comparison. We present results on \texttt{LLaMA-3.1-8B} in Table~\ref{tab:llama3_results_full}, and additionally validate our findings on \texttt{LLaMA-2-7B} in Table~\ref{tab:ablation_blob} to demonstrate generalization across model architectures.

\begin{table}[h]
\centering
\caption{Ablation study comparing PoLAR-VBLL with PoLAR-based baselines on WG-S using \texttt{Llama-2-7B}. All methods use identical PoLAR adapters, isolating the contribution of different uncertainty quantification approaches.}
\label{tab:ablation_blob}
\begin{tabular}{c|c|c|c}
\toprule
\textbf{Method} & \textbf{ACC (\%)} $\uparrow$ & \textbf{ECE (\%)} $\downarrow$ & \textbf{NLL} $\downarrow$ \\
\midrule
PoLAR-LA & 70.33$\pm$0.69 & 12.16$\pm$2.58 & \underline{0.69$\pm$0.03} \\
PoLAR-LA-LL & 70.33$\pm$0.69 & 14.63$\pm$1.14 & 0.71$\pm$0.05 \\
PoLAR-BLoB & \underline{70.39$\pm$0.26} & \underline{12.06$\pm$0.81} & 0.73$\pm$0.04 \\
\midrule
PoLAR-VBLL (w/o LA) & \textbf{71.62$\pm$0.27} & 8.26$\pm$0.60 & 0.66$\pm$0.03 \\
\textbf{PoLAR-VBLL (Full)} & \textbf{71.62$\pm$0.27} & \textbf{7.31$\pm$0.32} & \textbf{0.60$\pm$0.01} \\
\bottomrule
\end{tabular}%
\end{table}

\paragraph{VBLL is the Primary Driver of Calibration, Not LA.}
As shown in both Table~\ref{tab:llama3_results_full} and Table~\ref{tab:ablation_blob}, PoLAR-VBLL (w/o LA) already achieves strong calibration performance across datasets without any Laplace refinement. The subsequent LA step provides consistent but incremental improvements in ECE and NLL. This demonstrates that VBLL constitutes the core working mechanism for uncertainty quantification, with LA serving as a complementary refinement rather than a remedial component.

\paragraph{VBLL is the Primary Driver of Calibration, Not PoLAR.}
All variants in Table~\ref{tab:ablation_blob} employ identical PoLAR adapter structures, yet their calibration performance varies dramatically. PoLAR-LA, PoLAR-LA-LL, and PoLAR-BLoB achieve comparable but relatively poor ECE, while PoLAR-VBLL (w/o LA) delivers substantially better calibration. This confirms that the performance improvements are attributable to the variational training framework rather than the adapter architecture alone.

\paragraph{Limitations of Deterministic Training with Post-hoc LA.}
The comparison between PoLAR-LA and PoLAR-LA-LL in Table~\ref{tab:llama3_results_full} reveals that applying LA exclusively to the last layer of a deterministically trained model yields substantially worse calibration than applying LA across all adapter layers. This performance gap indicates that deterministic training fails to discover posterior geometries amenable to uncertainty quantification, necessitating LA compensation across all layers to achieve reasonable calibration.

\paragraph{VBLL Discovers High-Quality Posterior Modes During Training.}
A striking observation emerges from comparing PoLAR-VBLL to PoLAR-LA: despite applying LA only to the last layer, PoLAR-VBLL achieves superior calibration, whereas PoLAR-LA applies LA across all adapter layers. This demonstrates that the strong calibration stems from VBLL's variational training, not from LA correcting a deficient posterior. VBLL actively guides optimization toward high-quality posterior modes that inherently support reliable uncertainty estimation.

\subsection{Tightness of the Jensen Bound}
\label{sec:jensen_bound_tightness}

A potential concern regarding our VBLL formulation is whether the Jensen bound employed in Eq.~\ref{eq:jensen_bound} provides a sufficiently tight approximation to the true ELBO objective. To address this concern, we conduct an empirical comparison between our analytical Jensen-based estimator and a Monte Carlo (MC) estimator with 50 samples across the full training horizon.

Specifically, we train PoLAR-VBLL on the WG-S dataset using \texttt{Llama-3.1-8B} as the backbone and record the training loss computed by both estimators at regular intervals over 400 training steps. The results are presented in Table~\ref{tab:jensen_vs_mc}.

\begin{table}[h]
\centering
\caption{Comparison of training loss trajectories between the Jensen bound and 50-sample Monte Carlo estimation on WGS dataset using \texttt{Llama-3.1-8B}.}
\label{tab:jensen_vs_mc}
\begin{tabular}{c|ccc}
\toprule
\textbf{Training steps} & \textbf{VBLL (Jensen)} & \textbf{VBLL (50-sample MC)} & \textbf{Absolute Gap} \\
\midrule
0 & 69.50 & 61.23 & 8.27 \\
50 & 50.71 & 50.97 & 0.26 \\
100 & 45.57 & 45.65 & 0.08 \\
150 & 40.95 & 40.86 & 0.09 \\
200 & 36.96 & 36.95 & 0.01 \\
250 & 33.57 & 33.74 & 0.17 \\
300 & 31.08 & 31.36 & 0.28 \\
350 & 29.55 & 29.89 & 0.34 \\
400 & 28.49 & 28.83 & 0.34 \\
\bottomrule
\end{tabular}
\end{table}

The results reveal several important observations regarding the fidelity of our Jensen-based optimization. First, the initial gap between the two estimators (8.27 at step 0) undergoes rapid convergence within the first 50 training steps, decreasing to merely 0.26. This rapid alignment indicates that the Jensen bound quickly becomes an accurate proxy for the true objective as the model parameters move away from their random initialization.

Second, after this initial convergence phase, the absolute gap remains remarkably stable throughout the remainder of training, consistently staying below 0.35 from step 50 to step 400. This stability demonstrates that the Jensen bound maintains its approximation quality across the entire optimization trajectory, rather than degrading as the posterior distribution evolves during training.

Third, and most critically, the gap does not exhibit any increasing trend as training progresses. This absence of divergence confirms that optimizing the Jensen-based lower bound does not lead the model toward regions where the bound becomes loose or misleading. Instead, the Jensen estimator and the MC estimator track each other closely throughout the full training horizon.

These extended results provide strong empirical evidence that our analytical Jensen-based formulation maintains fidelity to the true ELBO objective. The tight correspondence between the two estimators validates our design choice of employing the Jensen bound, which enables efficient closed-form gradient computation without sacrificing optimization quality. This computational advantage is substantial: while the 50-sample MC estimator requires 50 forward passes through the last layer per training step, our Jensen-based approach achieves comparable optimization trajectories with a single analytical computation.

\subsection{Sensitivity to prior and initialization}
\label{app:sensitivity}

We conduct a comprehensive sensitivity analysis to evaluate the robustness of our method with respect to two critical factors: (1) the choice of prior distribution, and (2) the initialization of variational parameters. We investigate the sensitivity to the prior scale parameter $\sigma_\theta$, which controls the width of the Gaussian prior over the last-layer weights $\mathcal{N}(\mathbf{0}, \sigma_\theta^2 \mathbf{I})$. To assess initialization robustness, all experiments are conducted across three different random seeds $\{1, 2, 3\}$, which affect both data shuffling and stochastic aspects of variational parameter initialization. We report the mean and standard deviation across these seeds.

\begin{table}[h]
\centering
\caption{Sensitivity analysis of prior scale $\sigma_\theta$ on WG-S under LLaMA 2 7B. Results are averaged over three random seeds with standard deviations reported. ACC: accuracy (\%), ECE: expected calibration error (\%), NLL: negative log-likelihood.}
\label{tab:prior_sensitivity}
\begin{tabular}{c|ccc}
\toprule
Prior Scale ($\sigma_\theta$) & ACC ($\uparrow$) & ECE ($\downarrow$) & NLL ($\downarrow$) \\
\midrule
0.1 & $70.92 \pm 0.24$ & $9.10 \pm 0.53$ & $0.68 \pm 0.04$ \\
\textbf{1.0} (Default) & $\mathbf{71.62 \pm 0.27}$ & $\mathbf{8.26 \pm 0.60}$ & $\mathbf{0.66 \pm 0.03}$ \\
10.0 & $70.70 \pm 0.50$ & $10.59 \pm 1.17$ & $0.69 \pm 0.07$ \\
\bottomrule
\end{tabular}
\end{table}

Table~\ref{tab:prior_sensitivity} summarizes the performance under different prior scales on the WG-S dataset. We can make the following observations:

\textbf{(1) Optimal prior scale:} The prior scale $\sigma_\theta = 1.0$ achieves the best overall performance across all metrics. Both overly restrictive ($\sigma_\theta = 0.1$) and overly diffuse ($\sigma_\theta = 10.0$) priors result in degraded performance, with decreases in accuracy and increases in both calibration error and negative log-likelihood.

\textbf{(2) Convergence dynamics:} We observe that as the prior scale increases from 0.1 to 10.0, the optimization process converges progressively more slowly during training. This suggests that excessively wide priors introduce additional optimization challenges, potentially requiring more iterations to reach comparable solution quality.

\textbf{(3) Initialization robustness:} The relatively small standard deviations across different random seeds demonstrate that our method exhibits strong robustness to initialization. This stability is consistent across all tested prior scales, indicating that the variational learning process reliably converges to high-quality solutions despite variations in random initialization. The consistency across seeds also validates the reproducibility of our approach.


\subsection{Stable Rank Analysis and Theoretical Justification}
\label{sec:stable_rank_analysis}

In this section, we provide both theoretical motivation and empirical validation for combining PoLAR with VBLL. We first establish the theoretical foundation linking feature geometry to uncertainty quantification quality, and then present empirical evidence demonstrating that PoLAR preserves the geometric properties essential for reliable Bayesian inference.

\subsubsection{Theoretical Motivation: Distance-Aware Features for Bayesian Last Layer Methods}

Recent work on deterministic uncertainty quantification has established that last-layer Bayesian methods critically depend on the geometric properties of the feature extractor. In particular, SNGP~\citep{liu2020simple} demonstrates that distance-aware features, where semantically distinct inputs remain well-separated in the feature space, are essential for reliable uncertainty estimation. We argue that VBLL shares this requirement: when the Bayesian last layer receives features from a distance-preserving extractor, it can effectively distinguish between in-distribution (ID) and out-of-distribution (OOD) samples based on their relative positions in the feature space.

A critical failure mode arises when the learned transformation exhibits low effective dimensionality, a phenomenon termed feature collapse~\citep{postels2021practicality}. Under feature collapse, the adapter projects high-dimensional inputs onto a narrow, low-dimensional subspace, causing semantically diverse inputs, including OOD samples,to cluster together indistinguishably from ID data. This geometric compression fundamentally limits the Bayesian last layer's capacity to detect distribution shift, as the distance information necessary for uncertainty-aware inference is lost during feature extraction.

The stable rank of the learned weight update $\Delta \mathbf{W}$ provides a quantitative measure of this geometric property. Defined as
\begin{equation}
    \text{stable-rank}(\Delta \mathbf{W}) = \frac{\|\Delta \mathbf{W}\|_F^2}{\|\Delta \mathbf{W}\|_2^2},
    \label{eq:stable_rank}
\end{equation}
the stable rank captures the effective dimensionality of the transformation by computing the ratio of the squared Frobenius norm to the squared spectral norm. A stable rank approaching 1.0 indicates a nearly rank-1 projection that severely compresses the feature space, while higher values suggest a more isotropic transformation that preserves multiple effective directions.

\subsubsection{Empirical Validation: PoLAR Preserves Feature Geometry}

\begin{figure}[t]
    \centering
    \includegraphics[width=1.0\linewidth]{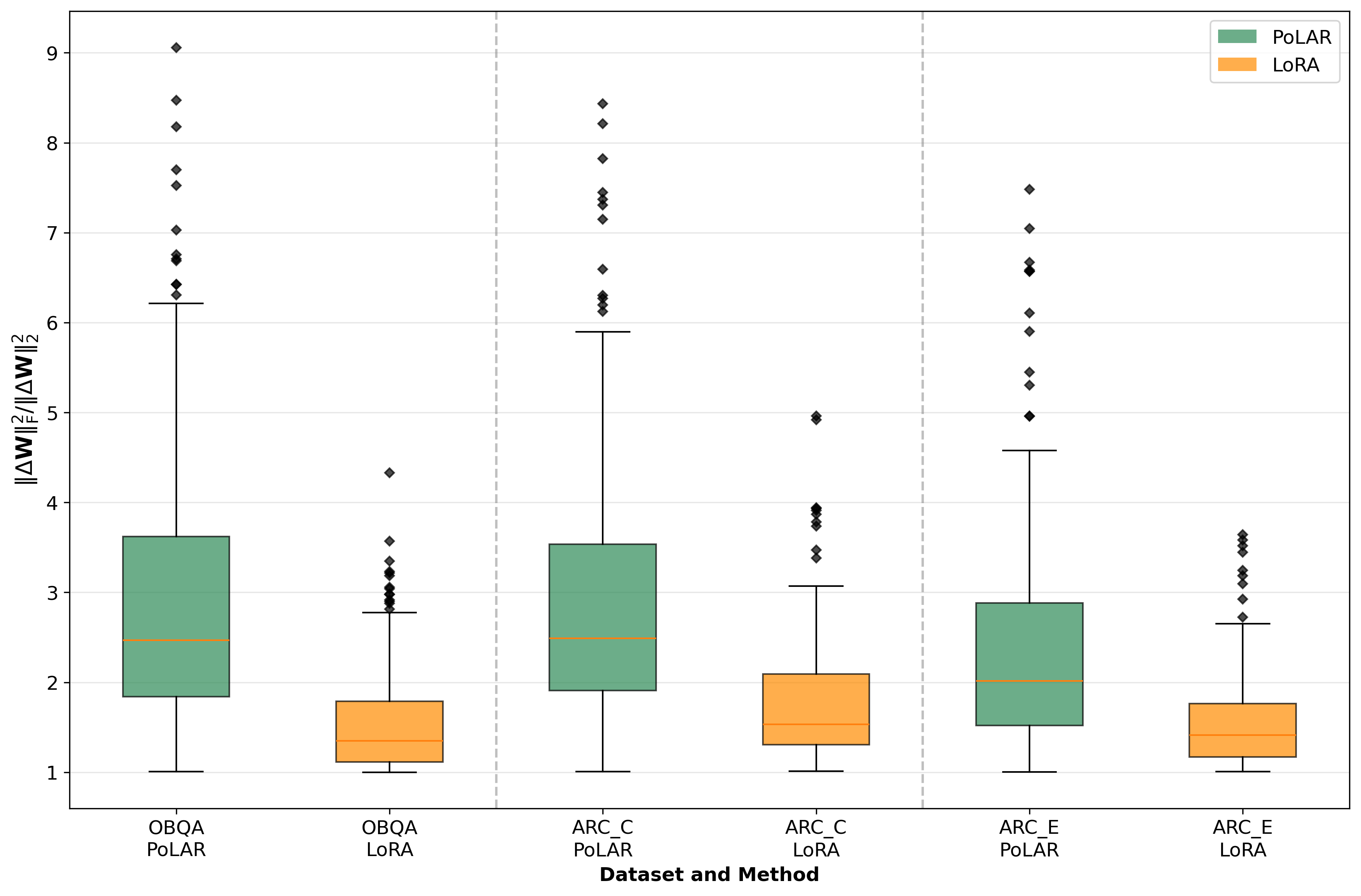}
    \caption{Stable rank comparison between PoLAR and LoRA across three datasets. PoLAR consistently achieves higher stable rank values, indicating better preservation of feature geometry and effective utilization of the allocated parameter space.}
    \label{fig:stable_rank_analysis}
\end{figure}

To empirically validate our theoretical motivation, we conduct a comparative stable rank analysis between LoRA and PoLAR across multiple datasets. Figure~\ref{fig:stable_rank_analysis} presents the distribution of stable rank values for both methods.

The results reveal a striking contrast between the two adaptation strategies. Standard LoRA exhibits an average stable rank of approximately 1.53, approaching the theoretical minimum of 1.0. This low value indicates that LoRA effectively performs a nearly rank-1 projection, compressing the learned updates into a highly anisotropic subspace despite the nominally higher allocated rank. Such geometric compression aligns with previous observations of rank collapse in LoRA~\citep{zhang2025polar} and explains the suboptimal performance of LoRA-based uncertainty quantification methods, particularly in OOD detection scenarios where distance preservation is critical.

In contrast, PoLAR maintains a significantly higher average stable rank of approximately 2.86. By constraining the low-rank factors $\mathbf{U}$ and $\mathbf{V}$ to the Stiefel manifold through orthogonality constraints, PoLAR encourages a more isotropic transformation that preserves multiple effective directions in the feature space. This geometric preservation directly supports the requirements of VBLL: the Bayesian last layer receives features that maintain semantic distances between inputs, enabling more reliable uncertainty estimation for both ID and OOD samples.

The connection between stable rank and downstream performance is evident in our experimental results. Across all evaluation benchmarks, PoLAR-based methods consistently outperform their LoRA counterparts in both predictive accuracy and uncertainty calibration. The higher stable rank of PoLAR translates to richer feature representations that better capture task-specific patterns while preserving the geometric structure necessary for principled Bayesian inference.

\end{document}